\pdfoutput=1

\documentclass[11pt]{article}

\usepackage{supertabular,cuted}
\usepackage{acl}

\usepackage{times}
\usepackage{latexsym}
\usepackage{multirow}
\usepackage{makecell}
\usepackage[T1]{fontenc}

\usepackage[utf8]{inputenc}

\usepackage{microtype}

\usepackage{graphicx}
\usepackage{booktabs}
\usepackage{lipsum}  
\usepackage{tabularx}
\usepackage{booktabs}
\usepackage{microtype}
\usepackage{type1cm}
\usepackage[american]{babel}
\usepackage{amssymb}
\usepackage{multirow}
\usepackage{xspace}
\DeclareMathAlphabet{\mathcalbf}{OMS}{pzc}{b}{n}
\usepackage{enumitem}
\usepackage{colortbl}
\usepackage{enumitem}
\usepackage{tablefootnote}
\RequirePackage{color}
\definecolor{darkgray}{gray}{0.40}
\definecolor{mediumgray}{gray}{0.60}
\definecolor{lightgray}{gray}{0.95}
\definecolor{ultralightgray}{gray}{0.98}
\definecolor{forestgreen}{rgb}{0.133, 0.545, 0.133}
\definecolor{orange}{rgb}{1, 0.86, 0.74}
\definecolor{lightergreen}{rgb}{0.95, 1, 0.88}

\usepackage{graphicx}
\DeclareGraphicsExtensions{.pdf,.ai,.jpg,.png}
\setkeys{Gin}{pagebox=artbox}
\graphicspath{{./figures/}}

\RequirePackage{type1cm}
\RequirePackage{color}
\RequirePackage{soul}
\setstcolor{blue}
\definecolor{violet}{rgb}{0.5,0.0,0.5}

\newcommand{\ls}{\addlinespace[3.5pt]}

\newsavebox\bscombox
\newcommand{\bscom}[3][]{%
    \sbox{\bscombox}{\fontsize{8}{9}\selectfont#1#2#3}
    \noindent
    \st{#2}{\selectfont
        \color{blue}#3\ifx\\#1\\\else{\fontsize{8}{9}\selectfont\color{violet}[#1]}\fi
    }
}

\begin{document}
\title{ArgBench: Benchmarking LLMs on Computational Argumentation Tasks}


\author{
	Yamen Ajjour\\
	Leibniz University Hannover \\
	{\tt y.ajjour@ai.uni-hannover.de}\\\And
	\hspace{2em}Carlotta Quensel \\
	\hspace{2em}Leibniz University Hannover \\
	\hspace{2em}{\tt c.quensel@ai.uni-hannover.de} \\\AND
	Nedim Lipka \\
	Adobe Research \\
	{\tt lipka@adobe.com} \\\And
	\hspace{2em}Henning Wachsmuth\\
	\hspace{2em}Leibniz University Hannover, L3S \\
	\hspace{2em}{\tt h.wachsmuth@ai.uni-hannover.de}
}


\maketitle
\begin{abstract}
Argumentation skills are an essential toolkit for large language models (LLMs). These skills are crucial in various use cases, including self-reflection, debating collaboratively for diverse answers, and countering hate speech. In this paper, we create the first benchmark for a standardized evaluation of LLM-based approaches to computational argumentation, encompassing 33 datasets from previous work in unified form. Using the benchmark, we evaluate the generalizability of five LLM families across 46 computational argumentation tasks that cover mining arguments, assessing perspectives, assessing argument quality, reasoning about arguments, and generating arguments. On the benchmark, we conduct an extensive systematic analysis of the contribution of few-shot examples, reasoning steps, model size, and training skills to the performance of LLMs on the computational argumentation tasks in the benchmark.
\end{abstract}

\section{Introduction}

Utilizing the capabilities of large language models (LLMs) for argumentation has drawn significant interest within the NLP community. Some promising applications where argumentation is applied include countering hate speech \cite{saha:2023a}, rebutting misinformation \cite{zheng:2025}, and suggesting various arguments for the possible answers of an input task \cite{miandoab:2024}. Going beyond step-by-step reasoning to understand and generate conflicting reasoning paths helps enhance LLMs' ability to detect reasoning gaps and effectively counter societal harms. This requires not only effective LLM-based methods but also comprehensive benchmarks for computational argumentation.

In general, research on computational argumentation is motivated by applications such as debating machines \cite{slonim:2021} and writing support \cite{skitalinskaya:2021}. Fundamental \nopagebreak{\emph{skills} needed} for these endeavors include mining arguments \cite{stab:2017a}, assessing their perspectives \cite{barhaim:2017} and their quality \cite{wachsmuth:2017b} as well as classifying reasoning patterns of arguments \cite{lawrence:2016} or generating counterarguments to input arguments \cite{hua:2019}.

While computational argumentation can equip LLMs with human-like reasoning skills \cite{stahl:2025}, existing benchmarks for tracking LLM progress largely overlook respective tasks. Reasoning benchmarks like ARC \cite{clark:2018} or MMLU \cite{hendrycks:2020} are limited to tasks with one correct answer, such as mathematics and common sense reasoning. Argumentation, however, deals with questions on complex topics, with conflicting perspectives and divergent reasoning paths, such as ``Should we increase the minimum wage?''. The few existing argumentation-specific LLM works are limited in scope (details in Section~\ref{sec:relatedwork}), covering only few skills, such as argument mining and generation \citep{chen:2024}.

In this paper, we present a benchmark to assess the capabilities of LLMs over 33 computational argumentation datasets. The benchmark includes 46~tasks that are grouped into five skills: argument mining, argument quality assessment, argument perspective assessment, argument reasoning, and argument generation. It offers a comprehensive evaluation framework for LLMs, covering a broad range of argumentation tasks. The benchmark provides two setups to evaluate LLMs: \emph{prompting} and \emph{leave-one-task-out} per skill. In the prompting setup, an LLM is primed with the task definition, without fine-tuning. In leave-one-task-out, we evaluate the model's ability to generalize to unseen tasks. Specifically, an LLM is evaluated on one task from each skill after fine-tuning it on the remaining tasks in the benchmark. The two evaluation setups assess the computational argumentation skills that a model possesses and its ability to generalize to new computational argumentation tasks. 

Based on the new benchmark, we evaluate five LLM families that capture different architectures and post-training methods. Our experiments in prompting indicate reasonable performance for some skills (an F$_1$-score of 0.486 in perspective assessment for few-shot prompting, on average), while still leaving notable room for improvement in general. Furthermore, LLMs' cross-task generalization appears limited, though models demonstrate better generalizability on specific argument reasoning and quality assessment tasks.

The main contributions of the paper are: 
\begin{itemize}
	\setlength{\itemsep}{0pt}
	\item 
	A benchmark to develop and assess the computational argumentation skills of LLMs covering 33 datasets and 46 tasks.
	\item A quantitative analysis of LLMs' generalization across these tasks.%
	\item 
	Extensive insights into which skills LLMs are good at by default, and which require more resources or more sophisticated approaches.%
	\footnote{Our experiment code can be accessed here: \url{https://github.com/webis-de/argbench}}
\end{itemize}









\section{Related Work}
\label{sec:relatedwork}

Existing training and evaluation methodologies for LLMs show that argumentation enhances LLMs' capabilities on different tasks. Among these, \citet{du:2024} train LLMs to debate on six tasks that encompass factuality and mathematical reasoning. On all tasks, the debating models outperform models on their own, showing the merit of multi-agent debating for factuality and reasoning. \citet{liang:2024} similarly demonstrate the advantage of multi-agent debate over self-reflection for machine translation and unintuitive arithmetic reasoning . In contrast, \citet{lin:2024} stress that the ability of LLMs to self-criticize is crucial for logic-centric tasks, such as code generation and arithmetic reasoning. \citet{miandoab:2024} compare argument generation to chain-of-thought prompting by letting an LLM generate arguments for candidate answers and then picking the answer with the best argument. They demonstrate the benefit of argument generation on bias detection datasets, such as StereoSet \cite{nadeem:2021} and DiFair \cite{zakizadeh:2023}. \citet{khan:2024} advise an approach where two LLMs debate to convince an LLM-based judge with the right answer in reading comprehension. The experiments show that even weaker judges can improve the ability of stronger models in reading comprehension. Altogether, prior works thus underline the potential of systematically evolving LLMs toward computational argumentation skills, for which we lay the ground in this paper.

Computational argumentation studies skills such as mining arguments from natural language text \cite{stab:2017a}, assessing their perspectives \cite{barhaim:2020}, quality \cite{wachsmuth:2017b}, and reasoning \cite{habernal:2018a} as well as generating arguments \cite{hua:2019}. Existing benchmarks for computational argumentation focus on single or a few tasks. \citet{feger:2025} combine multiple argument mining datasets to systematically test the models' generalization capacities, showing their limited generalizability over argument mining datasets. A similar benchmark is proposed by \citet{gemechu:2024} for argument relation identification. Beyond their restriction to argument mining, neither of benchmarks targets LLMs nor includes an LLM in their evaluation. 

The landscape of benchmarks for LLMs is also focused on single or a few tasks~\cite{pietron:2025, abkenar:2024}. \citet{gurjar:2025} evaluate LLMs at mapping arguments in long discussions to key points. A similar benchmark in terms of focus is \citet{dhole:2025} which uses LLM judges to evaluate retrieval-augmented argument generation. The most comprehensive benchmark of LLMs in terms of genre is proposed by \citet{gemechu:2025}, which is focused on the task of identifying a missing premise or conclusion in an argumentation in seven genres. Compared to these benchmarks, the breadth of tasks in our benchmark largely exceeds any existing study. Moreover, we design a setting that evaluates LLMs' ability to generalize to unseen tasks and enables the use of model-independent measures (noting reliability limits for argument generation).

Similar to benchmark studies, LLM-based approaches to computational argumentation are still few, mostly focusing on single tasks or specific genres. \citet{lin:2023} explore instruction fine-tuned LLMs for argument generation, finding that they largely outperform standard models on argumentation datasets. \citet{chen:2024} investigate the effectiveness of LLMs at argument mining and generation, while \citet{deshpande:2024} study to what extent LLM-generated context information benefits quality assessment, and \citet{mouchel:2025} align the argument generation skills of an LLM towards reasoning skills. In recent work, \newcite{stahl:2025} develop a specialized instruction fine-tuning process to generalize LLMs' capabilities across computational argumentation tasks. While they release the seed tasks used in their self-instruct \citep{wang:2023} pipeline, this data does not cover the full spectrum of tasks. As such, all
these approaches may serve as starting points for different aspects of our benchmark, which covers 46 computational argumentation tasks in total.

\section{The ArgBench Benchmark Dataset}

\renewcommand*{\thefootnote}{\fnsymbol{footnote}}
\begin{table*}[!t]
	\small
	\renewcommand{\arraystretch}{0.955}
	\setlength{\tabcolsep}{2pt}
	\begin{tabular}{@{}llll@{}}
		\toprule
		\bf Task & \bf Input & \bf Output & \bf Source \\
		\midrule
		
		\multicolumn{4}{@{}l@{}}{\textbf{Argument Mining}} \\    
		Conclusion Extraction & Sentence\footnotemark[1]  & \texttt{conclusion} or \texttt{no-conc.} & \citet{poudyal:2020}\\
		
		Premise Extraction & Sentence\footnotemark[1] & \texttt{premise} or \texttt{no-premise} & \citet{poudyal:2020}\\
		Premise Extraction & Motion, Evidence & \texttt{accept} or \texttt{reject} & \citet{eindor:2020}\\    
		Relation Detection & Sentence pair\footnotemark[1] & \texttt{related} or \texttt{not-related} & \citet{poudyal:2020}\\
		
		Relation Identification & Unit pair\footnotemark[1] & \texttt{attack} or \texttt{support} & \citet{peldszus:2014}\\
		Relation Identification & Unit pair\footnotemark[1] & \texttt{attack} or \texttt{support} & \citet{stab:2017a}\\
		Relation Identification & Unit pair\footnotemark[1] & \texttt{attack} or \texttt{support} & \citet{skeppstedt:2018}\\            
		Relation Identification & Argument pair & 3 Labels (e.g., \texttt{attack}) & \citet{menini:2018}\\
		Relation Identification & Unit pair\footnotemark[1] & \texttt{evidence} or \texttt{reason} & \citet{park:2018} \\
		
		Unit Classification & Argument unit\footnotemark[1] & 3 Labels (e.g., \texttt{claim}) & \citet{stab:2017a} \\
		Unit Classification & Argument unit\footnotemark[1] & 6 Labels (e.g., \texttt{anecdote}) & \citet{alkhatib:2017}\\
		Unit Classification & Argument unit\footnotemark[1] & 5 Labels (e.g., \texttt{fact}) & \citet{park:2018}\\    
		Unit Segmentation & Document & List of (\texttt{arg.} or \texttt{non-arg.}, Span)  & \citet{habernal:2015}\\
		Unit Segmentation & Document &List of (\texttt{arg.} or \texttt{non-arg.}, Span) & \citet{alkhatib:2016} \\
		Unit Segmentation & Document & List of (\texttt{arg.} or \texttt{non-arg.}, Span)& \citet{stab:2017a}\\
		Unit Segmentation & Document &List of (\texttt{arg.} or \texttt{non-arg.}, Span)  & \citet{poudyal:2020}\\
		
		\midrule
		\multicolumn{4}{@{}l@{}}{\textbf{Argument Perspective Assessment}}\\
		Argument Similarity & Argument pair & 4 Labels (e.g., \texttt{dissimilar}) &\citet{reimers:2019} \\
		Aspect Detection & Argument unit &   List of (\texttt{aspect} or \texttt{no-asp.}, Span)  & \citet{schiller:2021}\\
		Frame Identification & Argument, Frame & \texttt{match} or \texttt{no-match} & \citet{ajjour:2019}\\    
		Key Point Matching & Argument, Key Point & \texttt{match} or \texttt{no-match} & \citet{barhaim:2020} \\
		Stance classification & Topic, Conclusion & \texttt{pro} or \texttt{con} & \citet{barhaim:2017}\\
		Stance classification & Topic, Conclusion & \texttt{pro} or \texttt{con} & \citet{stab:2018}\\
		
		\midrule
		\multicolumn{4}{@{}l@{}}{\textbf{Argument Quality Assessment}}\\    
		Argument Ranking & Premise pair & \texttt{better} or \texttt{worse} & \citet{gretz:2020}\\
		Argument Ranking & Argument pair & \texttt{better} or \texttt{worse} & \citet{gleize:2019}\\
		Argument Ranking & Conclusion pair & \texttt{better} or \texttt{worse} &\citet{skitalinskaya:2021}\\
		Argument Rating & Argument & \texttt{Low}, \texttt{medium}, or \texttt{high} & \citet{wachsmuth:2017b}\\
		Argument Rating & Argument & 6 Labels (e.g., \texttt{very high}) &\citet{ng:2020}\\
		Controversy Scoring & Argument & 3 Labels (e.g., \texttt{very contr.}) & \citet{habernal:2018b}\\
		Reasonableness Scoring & Argument & 3 Labels (e.g., \texttt{neutral}) & \citet{habernal:2018b}\\
		Suboptimal Claim Det. & {Claim} & \texttt{improvable} or \texttt{not-imp.} & \citet{skitalinskaya:2023a}\\
		Claim Improve. Sugg. & {Claim, Quality Issue} & \texttt{match} or \texttt{no-match} & \citet{skitalinskaya:2023a}\\
		
		\midrule
		\multicolumn{4}{@{}l@{}}{\textbf{Argument Reasoning}}\\        
		Ad-hominem Detection & Argument & \texttt{ad-hominem} or \texttt{not-ad-hom.} & \citet{habernal:2018b}\\
		Fallacy Detection & Conclusion & 13 Labels (e.g., \texttt{ad-hominem})& \citet{jin:2022}\\
		Fallacy Detection & Argument unit &  List of (7 Fallacies , Spans) & \citet{goffredo:2023}\\
		Scheme Classification & Argument & 7 Labels (e.g., \texttt{false cause}) & \citet{saha:2023b}\\
		Warrant Identification & Claim, Two Warrants& \texttt{warrant 1} or \texttt{warrant 2} & \citet{habernal:2018a}\\
		
		\midrule    
		\multicolumn{4}{@{}l@{}}{\textbf{Argument Generation}}\\
		Argument Generation & Topic, Stance, Facts & Argument & \citet{saha:2023b}\\
		Argument Summariz. & Argument & Topic & \citet{roush:2024}\\
		
		Claim Generation & Topic Stance pairs & Argument & \citet{alshomary:2021}\\
		Claim Optimization & Conclusion & Conclusion & \citet{skitalinskaya:2023b}\\
		Counterargument Gen. & Argument & Argument & \citet{hua:2018}\\
		Counterargument Gen. & Argument & Argument & \citet{hua:2019}\\
		
		Warrant Generation & Conclusion, Premise & Warrant & \citet{bhagavatula:2020}\\
		Warrant Generation & Conclusion, Premise & Warrant & \citet{habernal:2018a}\\
		Warrant Generation & Conclusion, Premise & Warrant & \citet{boltuzic:2016}\\
		Warrant Generation & Conclusion, Premise & Warrant & \citet{becker:2020}\\
		\bottomrule
	\end{tabular}
	\caption{The 46 computational argumentation tasks in our benchmark, with modeled inputs and output. The output of classification tasks are labels with 1--2 words and encode a class (e.g., \texttt{pro} or \texttt{con}). For segmentation tasks, the output is a list of labels and the labeled text spans. For generation tasks, the output is a text span that represents an argumentative concept (e.g., Conclusion). $^*$We prepend the input with a context window that contains the input.}
	\label{table-modeled-tasks}
\end{table*}
\renewcommand*{\thefootnote}{\arabic{footnote}}

A key advantage of training LLMs at large scale to follow instructions is the generalization to unseen tasks enabled thereby \cite{ouyang:2022}. The generalization capabilities of LLMs are also crucial in computational argumentation, given the richness of aspects studied in this research area and the wide spectrum of genres that argumentation is relevant to. Hence, assessing the performance of LLMs on computational argumentation tasks requires a representative set of tasks.

To this end, we resort to the survey of \citet{lauscher:2022}, which groups tasks into four skills: argument mining, argument assessment, argument reasoning, and argument generation. However, we decided to divide  assessment into argument perspective assessment and argument quality assessment, since these skills capture conceptually-different aspects of argumentation. Based on this organization, we collected 46 tasks that cover many genres and a broad spectrum of how computational argumentation tasks are defined. Most of these come from the time before the LLM era. However, despite its different workings, we still expect a convincing LLM to be able to tackle them effectively.%
\footnote{The full benchmark can be found here:  \url{https://github.com/webis-de/argbench-data}}

Table~\ref{table-modeled-tasks} lists the 46 benchmark tasks we derived from 33 datasets. More details about the datasets, their sizes, and their covered nine genres are listed in Table~\ref{table-datasets} (Appendix). For all datasets, we uniformly defined the tasks as prompts that an LLM gets to complete with the appropriate answer. To ensure the quality of the benchmark, we employed consistent terminology, format, and instruction layouts and used the same evaluation measure for the same type of tasks. Whenever possible, we introduced unknown terminology as defined in the respective papers of the datasets. Table~\ref{table-modeled-tasks} outlines the input and output for each task. The task definitions as well as the input and output extraction and formatting can be found in Appendix~\ref{benchamrk-details}, with exemplary task definitions in Tables~\ref{table-prompts} to~\ref{table-prompts-last}.

According to how they are defined, the tasks fall into three groups: \emph{classification}, \emph{generation}, and \emph{segmentation}.  
In the segmentation tasks, the model splits a document into multiple spans with their labels. In argument unit segmentation, for example, a document is split into multiple spans which are prepended with \texttt{argumentative} or \texttt{non-argumentative}, indicating whether the span is part of an argument or not. In the classification tasks, the model labels an argument or an argument unit with one of multiple labels (e.g., \texttt{low} or \texttt{high} quality). In the generation tasks, the model generates a text segment (e.g., an argument) for a given input (e.g., a topic). 

Exemplarily, the prompt for the classification task \emph{suboptimal claim detection} \cite{skitalinskaya:2023a} looks as follows:

\begin{itemize}
\item[]
\tt Judge if the following claim can be improved by revising it. Possible outputs: \textbf{Improvable} if revision should be made, \textbf{Non-Improvable} if no revision is necessary. Only output Improvable or Non-Improvable.
\end{itemize}


\subsection{Evaluation Setup}

As an evaluation metric, we use macro F$_1$-score for all segmentation and classification tasks. For segmentation tasks, we measure the F$_1$-score for each positive class while considering a true positive in case of an exact match between an output span and a ground-truth span. We return the macro-average of all positive labels in case a task maps spans of text to multiple positive labels, for example argumentative in case of unit segmentation~\cite{ajjour:2017}. For generation tasks, we also need an automated metric despite its limitations. For meaningful insights, we utilize BertScore \cite{zhang:2020} to capture the semantic similarity between the ground-truth and generated text.

We adopt two evaluation setups that are tailored to existing learning methods for LLMs: \emph{prompting} and \emph{leave-one-task-out}. For both, we utilize the original test sets, when available. Otherwise, we split the dataset into training, validation, and test sets with a 60/20/20 ratio, guaranteeing no topic overlap between the sets to foster generalization across topics. As the size of the datasets is heavily skewed, we sample 1000 instances from validation and test sets whose size exceeds this threshold. Similarly, we sample 3000 instances from training sets whose size is larger than 3000.

\paragraph{Prompting} 

In the prompting setup, an LLM is evaluated on the test sets of the tasks without fine-tuning them. This setup evaluates an LLM's out-of-the-box computational argumentation abilities acquired during its training. The score of a skill in ArgBench is the macro-average of the scores of all the tasks under the skill. The size of the test sets for all tasks is 33,795 instances. Table~\ref{table-prompting-set-statistics} (Appendix) lists the number of instances for each skill.

\paragraph{Leave-One-Task-out} 

Here, we fine-tune the LLM on the training sets of all tasks except one, on which we evaluate the model. This setting assesses the ability of the LLM to learn computational argumentation skills and to generalize to unseen tasks. As unseen target tasks, we randomly sampled one task from each skill. The target tasks are: \emph{unit segmentation} on Web Discourse \cite{habernal:2015}, \emph{argument rating} \cite{wachsmuth:2017b}, \emph{argument similarity} \cite{reimers:2019}, \emph{ad-hominem detection} \cite{habernal:2018b}, and \emph{counterargument generation} \cite{hua:2018}. The remaining tasks are used for training. As a validation task, we select stance classification \cite{stab:2018}.






\section{Experiments}

In this section, we evaluate the performance of state-of-the-art LLMs in both setups of the benchmark, \emph{prompting} and \emph{leave-one-task-out}. We start by introducing the LLMs used in the experiments and then report on the prompting techniques used for each setup, before we come to the results and a subsequent manual evaluation.

\paragraph{LLMs}

We cover a total of ten state-of-the-art open-weight instruction fine-tuned LLMs from five model families of different sizes as well as one closed-weight LLM for comparison:
\begin{itemize}
\setlength{\itemsep}{-2pt}
	\item \texttt{Mistral} \cite{jiang:2023} with 7 and 22 billion parameters, and \texttt{Mixtral-8x7b} \cite{jiang:2024} with 56 billion parameters
	\item \texttt{Qwen3} \cite{yang:2025} with  4 and 32 billion parameters
	\item \texttt{Llama-3.1/3.3} \cite{dubey:2024} with 8 and 70 billion parameters
	\item \texttt{DeepSeek-R1} \cite{deepseek:2025} with 7 and 32 billion parameters
	\item \texttt{Phi-3.5-MoE-7.6b} \cite{abdin:2024} with 7.6 billion parameters
	\item \texttt{GPT-4.1} \cite{openai:2024}, estimated to have 1.8 trillion parameters
\end{itemize}

\begin{table*}[!t]
	\centering
	\small
	\renewcommand{\arraystretch}{0.88}
	\setlength{\tabcolsep}{5pt}
	\begin{tabular}{@{}llcccccc@{}} 
		\toprule
		\textbf{LLM} & \textbf{Prompt} & \textbf{Mining}    & \textbf{Perspective}  & \textbf{Quality} & \textbf{Reasoning} & \textbf{Generation}& \textbf{Macro}\\ 
	\midrule

		{\texttt{DeepSeek-R1-7b}}  & Zero-shot &  0.306	& 0.329 & 0.383	& 0.226 & 0.789 & 	0.407	\\ 

									& Few-shot  &  0.308 & 0.334	& 0.395  & 0.248 & 0.791 &	 0.415	\\
									& Chain-of-thought &  0.373 & 0.422   & 0.389	& 0.270 &  0.814&  \textbf{0.454}	\\
		
		\ls

		{\texttt{DeepSeek-R1-32b}}  & Zero-shot        &  0.352			 & 0.362			& 0.373 			& 0.292  		& 0.820		&0.440\\ 							
		& Few-shot 		   &  0.333			 & 0.363			& 0.402 			& 0.286 		& 0.807 	&0.438\\
		& Chain-of-thought & 0.406  		 & 0.478 			& 0.397			    & 0.274 		& 0.832 	&\textbf{0.477}\\

		\ls
		{\texttt{Llama-3.1-8b}}  	  & Zero-shot 			&  0.388 		 & 0.486 		& 0.377 			& 0.387 		& 0.836 	 &\textbf{0.495}\\
										& Few-shot 			&  0.367		 & 0.465 		& 0.358				& 0.332 		& 0.834 	 &	0.471\\
										& Chain-of-thought 	&  0.396 		& 0.492 		& 0.367 			& 0.389 		& 0.827		 &	0.494\\
		\ls
		{\texttt{Llama-3.3-70b}}  & Zero-shot		 &  0.490   & 0.552	& 0.419 	& 0.524   	& 	0.836	& 0.564\\ 
							  	  & Few-shot 		 & 0.527	& 0.620 & 0.418		& 0.526 	&0.844    	& \textbf{0.587}\\
							  	  & Chain-of-thought & 0.460  	& 0.531 & 0.439		& 0.396		&0.819	  & 0.529\\
		\ls
		{\texttt{Mistral-7b}}  & Zero-shot  	   & 0.385 			&  0.331  & 0.360  				& 0.356 		& 0.832   	 & 0.453\\
				     			& Few-shot   	   & 0.434  		&  0.316  & 0.344  				& 0.353 		& 0.844   	 & 0.458\\
								& Chain-of-thought & 0.441  		&  0.388  & 0.404 				& 0.351 		& 0.831      & \textbf{0.483}\\
		\ls
		{\texttt{Mistral-22b}}  & Zero-shot 		& 0.431 		& 0.409   & 0.335 				& 0.345 		& 0.834 	& 0.471\\ 

								& Few-shot 			& 0.400 		&  0.463  & 0.311 				& 0.463		    & 0.847    	& \textbf{0.497}\\
								& Chain-of-thought 	& 0.400			& 0.474   & 0.353			    & 0.430			& 0.829  	& 0.497\\
		\ls
		{\texttt{Mixtral-8x7b}}  & Zero-shot 		& 0.427  		 & 0.556 & 0.332 				& 0.395			& 0.833 	& 0.508\\ 
		& Few-shot 			& 0.443		     & 0.605 & 0.390				& 0.398			& 0.837  	&\textbf{0.535}\\
		& Chain-of-thought 	& 0.409  		 & 0.523 & 0.356 				& 0.389 		& 0.829  	& 0.501\\

		\ls
		{\texttt{Qwen3-4b}}  	    & Zero-shot 		& 0.378  	& 0.476 & 0.332			 		& 0.336 		& 0.839		& 0.472\\
									& Few-shot 			& 0.406  	& 0.494 & 0.367 			    & 0.380 		& 0.847		& \textbf{0.499}\\
									& Chain-of-thought 	& 0.375		& 0.484	& 0.407					& 0.278 		& 0.809	   	& 0.471\\
		\ls
		{\texttt{Qwen3-32b}}& Zero-shot	& 0.415 		& 0.533 	& 0.394 & 0.484			 		& 0.837		    & 0.533\\ 
								  
							& Few-shot 			& 0.501  & 0.588 	& 0.422 & 0.495	  				& 0.843 	    &  \textbf{0.570}\\
							& Chain-of-thought 	& 0.450  & 0.552 	& 0.430	& 0.358 				& 0.820 		& 0.522\\
							
		\ls
		{\texttt{Phi-3.5-MoE-7.6b}} & Zero-shot 		& 0.385  			& 0.410 				& 0.321 	& 0.366			 & 0.825 	& 0.461\\		
		& Few-shot 			& 0.325  			& 0.429				 	& 0.329 	& 0.332 		 & 0.840	 	&0.451\\
		& Chain-of-thought 	& 0.399 			& 0.398 				& 0.358 	& 0.388 		& 0.818 	& \textbf{0.472}\\						
							
	\midrule
		{\texttt{GPT-4.1}}  & Zero-shot 	&  0.536 & 0.607	& 0.440 & 0.556 & 0.849 	 &	0.597	\\ 
		& Few-shot 	&  0.597 & 0.664	& 0.477 & 0.560 & 0.865 	 &	\textbf{0.633}	\\ 
		& Chain-of-thought & 0.542 & 0.603 & 0.436 & 0.465 & 0.840 & 0.577\\							
	\bottomrule
	\end{tabular}
	\caption{\textbf{Results for prompting setup.} Performance of all LLMs in zero-shot, few-shot, and chain-of-thought prompting, on all tasks of each of the five skills: argument \emph{mining}, \emph{perspective} and argument \emph{quality} assessment, argument \emph{reasoning}, and \emph{generation}. \emph{Macro} is the macro average over all skills (best macro per LLM bold).}
	\label{table-prompting-results}
\end{table*}

\noindent

We used \url{huggingface.co} to load all open-weight LLMs. For the prompting experiments, we load the models with \texttt{bfloat16} precision. The exact model versions can be found in Table~\ref{table-model-links} in the Appendix. To foster reproducibility, we run all experiments with a fixed seed: 1516.

\subsection{Prompting Setup}

LLMs enable dedicated prompting techniques for a given task, such as \emph{few-shot} and \emph{chain-of-thought} prompting. Their impact on computational argumentation tasks is largely unexplored. To close this gap, we employ respective prompting approaches tailored to the benchmark tasks:

\paragraph{Zero-Shot} 

For each task, we prompt the model with its definition followed by the instance. For exemplary prompts, see Tables~\ref{table-prompts}--\ref{table-prompts-last} (appendix).

\paragraph{Few-Shot}

We randomly select examples from the training set and append them to the task definition. We sample four examples in our experiments since the average label count in the classification tasks is 3.13. This allows one example per label on average.

\paragraph{Chain-of-Thought}

We expect chain-of-thought prompting \cite{wei:2022} to be a promising technique for computational argumentation where tasks often compose multiple steps by concept. For example, to generate a counterargument to an argument, concrete steps can include extracting the main conclusion and premise of the argument, attacking each of them separately, and finally synthesizing the rebuttal. To trigger chain-of-thought across tasks while clearly delineating the final output from the reasoning steps, we add the following instruction after the task description: ``Think step by step and prepend your output with Output:''

\subsection{Leave-One-Task-Out Setup}
\label{sec:leave-one-task-out}

For this setup, we fine-tuned LLMs with 4-bit precision using LoRa \cite{hu:2022} in either of two training settings:

\paragraph{Training on All Other Tasks}

We fine-tuned each open-weight LLM on all tasks except for the target task using the respective instructions.\footnote{For lack of resources, we excluded \texttt{Llama-3.3-70b} and \texttt{Mixtral-8x7b} in the leave-one-task-out-experiments.} Then, we evaluated it on the target task. On the validation task, we optimized the learning rate and early stopping threshold using the Tree-structured Parzen Estimator~\cite{watanabe:2023}. Details on hyperparameter optimization are in Appendix~\ref{sec:implementation}, including tested ranges (Table~\ref{table-hyper-parameter-range}) and the best values (Table~\ref{table-hyper-parameters-cols-bs-lr}).

\paragraph{Training on the Target Task}

To assess the generalizability of LLMs to unseen tasks, we evaluate LLMs on each target task, after fine-tuning them on the training set of that task. This gives us an upper bound on tackling the task with LLMs.


\begin{table*}[t]
	\small
	\centering
	\renewcommand{\arraystretch}{0.825}
	\setlength{\tabcolsep}{1.5pt}
	\begin{tabular}{llccccccc} 
		\toprule
		\textbf{Model} & \textbf{Training} & \textbf{Mining}    & \textbf{Perspective}  & \textbf{Quality} & \textbf{Reasoning} & \textbf{Generation}& \textbf{Macro}\\ 
		     \cmidrule(l@{2pt}r@{2pt}){3-3}  \cmidrule(l@{2pt}r@{2pt}){4-4}  \cmidrule(l@{2pt}r@{2pt}){5-5}  \cmidrule(l@{2pt}r@{2pt}){6-6}  \cmidrule(l@{2pt}r@{2pt}){7-7}
		 &  & \textbf{Unit}   & \textbf{Argument}  & \textbf{Argument} & \textbf{Ad-hominem} & \textbf{Counterargument}& \\ 
		 &  & \textbf{Segmentation}   & \textbf{Similarity}  & \textbf{Rating} & \textbf{Detection} & \textbf{Generation}& \\ 
		\midrule
		{\texttt{DeepSeek-R1-7b}}	&On the target task &	0.141			& 	0.490	 	& 0.512	 	& 	0.844	 & 0.797&	  0.557\\
									&On all other tasks & 	0.138			& 	0.462		& 0.519	 	& 	0.819	 & 0.784	&	 0.544\\
		\ls
		{\texttt{DeepSeek-R1-32b}}	&On the target task &	0.242			& 	0.491	 	& 0.513 	& 	0.903	 & 0.803	&		0.590\\
									&On all other tasks & 	0.207			& 	0.381		& \textbf{0.562}	& 	0.883 & 0.789& 0.564 \\
		
		\ls

		{\texttt{Llama-3.1-8b}}		&On the target task &	0.210			& 	0.469		& 	0.377	& 	0.893 & 0.800		& 0.550 \\
									&On all other tasks & 	0.159			& 	0.273		& 	0.419	& 	0.872 & 0.777 	& 0.500 \\
		\ls

		{\texttt{Mistral-7b}}		&On the target task	 & 	0.226			& 	0.483		& 0.542 	& 0.897 	& 0.821	& 0.590\\
								 	&On all other tasks  & 	0.191			& 	0.207		& 0.518	 	& 0.840 	& 0.784	&  0.508\\
		\ls

		{\texttt{Mistral-22b}}		&On the target task	 & 	\textbf{0.270}	& 	\textbf{0.636}&  	0.554	& 	\textbf{0.912}	 & \textbf{0.825}		& \textbf{0.639}	 \\
									&On all other tasks  & 	0.167			& 	0.468	  	&	0.529		& 	0.890	 & 0.791		& 0.569  \\
		
		\ls

		{\texttt{Qwen3-4b}}			&On the target task &       0.235		& 		0.480 	& 	0.381	 & 	0.876	 & 0.801		& 0.555 \\
									&On all other tasks & 		0.134		& 	 	0.447	& 	0.451	 & 	0.879	 & 0.789	& 0.540 \\
		\ls

		{\texttt{Qwen3-32b}}		&On the target task &	0.261			& 	0.474		& 	0.388 	& 	0.895	 & 	0.821&	 0.568 \\
									&On all other tasks & 	0.138			& 	0.248 		& 	0.280 	& 	0.803	 & 	0.781 & 0.450 \\
		\ls

		{\texttt{Phi-3.5-MoE-7.6b}}	&On the target task &	0.198			& 	0.445		& 	\textbf{0.562} 	& 	0.872	 &	0.791 & 0.574 \\
									&On all other tasks & 	0.173			& 	0.250 		& 	0.477			& 	0.818  & 0.781	 & 0.500 \\
		\midrule
		{\texttt{GPT-4.1}}          &No Training 	    &   0.212 			&   0.441 		& 0.504 & 0.867 & 0.821 & 0.569   \\
		\bottomrule
	\end{tabular}
	\caption{\textbf{Results for leave-one-task-out setup.} Performance of all evaluated LLMs on five target tasks \cite{habernal:2015,reimers:2019,wachsmuth:2017b,habernal:2018b,hua:2018}, when fine-tuning a LoRa adapter \emph{on the target task}'s training set or \emph{on all other tasks}' training sets. \emph{Macro} is the macro average over all tasks. The best result per column is marked bold.}
	
	\label{table-results-leave-one-task}
\end{table*}

\subsection{Results}

\paragraph{Prompting} 

Table~\ref{table-prompting-results} shows the results of all eleven LLMs for each of the five skills as well as their macro average. Generally, we see that GPT-4.1 performs best, but the largest open-weight models come rather close. In zero-shot prompting, the larger models of \texttt{Qwen3}, \texttt{Llama}, \texttt{Mistral}, and \texttt{DeepSeek-R1} outperform their smaller counterparts on the macro score. This suggests that model size benefits computational argumentation tasks by default. The gain of larger over smaller models is lowest for argument quality assessment with a drop of --0.025 for \texttt{Mistral} and a maximum of 0.062 for \texttt{Qwen3}. In contrast, increasing the model size achieves the highest gain on argument mining (0.037--0.102) and perspective assessment (0.033--0.078). When we increase the model parameters by factor 4 or more, LLMs achieve even higher performance on argument reasoning (0.066--0.148).

Few-shot prompting achieves the best macro-average score for large models (32 billion parameters or more), except \texttt{DeepSeek-R1-32b} which is optimized specifically for reasoning. Across all large models, few-shot prompting seems to be the most effective prompting technique for argument reasoning and generation. In contrast, it performs worse than other prompting techniques on argument quality assessment for the \texttt{Llama} and \texttt{Mistral} families. Our error analysis of the output of \texttt{Mistral-22b} in Appendix~\ref{sec:error-analysis} shows a tendency to prefer positive labels (e.g., \texttt{better} vs \texttt{worse}), especially on argument ranking tasks. 

For argument quality assessment, chain-of-thought outperforms zero-shot-prompting for all models except \texttt{Llama-3.1-8b} and \texttt{GPT-4.1}. Surprisingly, however, its performance remains similar to subpar to zero-shot prompting on argument reasoning tasks for the \texttt{Llama}, \texttt{Mistral}, and \texttt{Qwen3} families. Our error analysis of chain-of-thought in argument reasoning in Appendix~\ref{sec:error-analysis} reveals that the error source often lies in difficulties in processing emotional language and instruction following failures. Another challenge is the tendency of \texttt{Qwen3-32b} to enter unfinished reasoning chains (78\% instances of argument reasoning tasks). On argument mining, chain-of-thought is the most effective prompting technique for all small models (8 billion parameters or less), except \texttt{Qwen3-4b}. In contrast, few-shot prompting dominates other prompting techniques for large models on argument mining with the exception of \texttt{DeepSeek-R1-32b}. 

We conclude that argument quality assessment seems the most challenging argumentation skill for LLMs, matching results from related work \citep{stahl:2025}. This is demonstrated by the low performance of LLMs on this skill in general as well as the small gain from increasing the model size or applying prompting techniques. The low performance of applying chain-of-thought on argument reasoning for three model families signals the inadequacy of step-by-step reasoning techniques to characterize argument reasoning patterns and gaps.

\paragraph{Leave-One-Task-Out}

Table~\ref{table-results-leave-one-task} lists the results of eight open-weight LLMs on each target task when training on its training set (in-task) and when training on the training sets of all remaining 45 tasks (leave-one-task-out). For comparison, we also show GPT-4.1 without training.

\texttt{Mistral-22b} achieves the highest macro-average score over the five tasks (0.639), but F$_1$-scores of only 0.270 and 0.554 on unit segmentation \cite{habernal:2015} and argument rating \cite{wachsmuth:2017b} respectively. These results indicate that, even with fine-tuning on the target, certain basic computational argumentation tasks remain a challenge for LLMs. When trained on other tasks, \texttt{Mistral-22b} achieves the best performance on three target tasks, while \texttt{DeepSeek-R1-32b} performs the best with F$_1$-scores of 0.207 on unit segmentation and 0.562 on argument rating. On ad-hominem detection, both \texttt{DeepSeek-R1-32b} and \texttt{Mistral-22b} show only a drop of 0.02 from the in-task performance.

On argument rating, the \texttt{DeepSeek-R1} family achieves F$_1$-score gains of 0.005 to 0.049 points when utilizing training on other tasks compared to target-task training. This suggests the potential of some LLMs to transfer the skill of evaluating argument quality to novel tasks. In contrast, for unit segmentation, the results of all models are consistently lower when trained on other tasks, dropping by 0.003 to 0.123 points. A similar task transfer challenge can be observed in argument similarity (a drop of 0.028 to 0.276 points) and counterargument generation (decreases of 0.01 to 0.04 points).

Taken together, the experiments highlight the limited generalizability of LLMs across computational argumentation tasks. While models such as \texttt{DeepSeek-R1-32b} and \texttt{Mistral-22b} generalize well to ad-hominem detection and argument rating, their performance remains lower than that of fine-tuning models on other target tasks, suggesting a limited ability to generalize to unseen computational argumentation tasks. Challenging skills for generalization, according to our experiments, are argument mining and generation, as well as perspective assessment. This hints at the ability of LLMs to generalize argumentation concepts to tasks related to evaluating and judging argumentation rather than generating arguments or extracting arguments and their perspectives.

\subsection{Manual Evaluation}
Automatic evaluation is of limited reliability for generation tasks. To gain further insights, we manually evaluated the output of the LLMs on the counterargument generation task of \cite{hua:2018}. This task aims at synthesizing counterarguments that properly challenge the arguments posted by authors on ChangeMyView by leveraging supporting information retrieved from Wikipedia to craft the arguments. We sampled 50 arguments from the test set of the task and generated counterarguments with four selected models (\texttt{Mistral-22b}, \texttt{Mistral-7b}, \texttt{Qwen3-32b}, and \texttt{Phi-3.5-MoE-7.6b}) in the leave-one-task-out setup. These were the top four models on the counterargument generation task, and they cover different model sizes, architectures, and regions of origin. The setup yielded $50 \cdot 2 \cdot 4 =$ $400$ input triples (comprising the argument, supporting information, and generated counterarguments). Three annotators from Upwork evaluated these triples on three dimensions using a 5-point Likert scale (where a higher value indicates higher quality): \textit{argumentativeness}, \textit{countering}, and \textit{relevance}. More information about the annotation process is in the Appendix~\ref{manual-evaluation-instructions}, including a screenshot of the annotation interface in Figure~\ref{fig:annotation-interface}.

\begin{table}[!t]
	\small
	\centering
	\renewcommand{\arraystretch}{0.825}
	\setlength{\tabcolsep}{3.5pt}
	\begin{tabular}{llrrrr} 
		\toprule
		\textbf{Model} & \textbf{Training} & \textbf{Arg}    & \textbf{Cou}  & \textbf{Rel} \\ 
		\midrule		
		{\texttt{Mistral-7b}}		&On the target task &       3.120		& 		3.120 	& 	2.533 	\\
									& On all other tasks & 		2.460		& 	 	2.753	& 	2.020 \\

		\ls
		{\texttt{Mistral-22b}}		&On the target task	 & 	\textbf{3.487}	& 	\textbf{3.493}	&  	\textbf{2.670} 	\\
									& On all other tasks  & 	2.680			& 	2.807	  	&	2.220 \\

		\ls
		{\texttt{Qwen3-32b}}		&On the target task &	3.227			& 	3.187		& 	2.647 \\
									& On all other tasks & 	2.513			& 	2.740 		& 	2.087  \\
		\ls							%

		{\texttt{Phi-3.5-MoE}}	& On the target task &	2.253			& 	2.473		& 	1.980 \\
		{\,\,\texttt{-7.6b}}							&On all other tasks & 	2.127			& 	2.433 		& 	2.027 \\
		\bottomrule
	\end{tabular}
	\caption{\textbf{Manual evaluation results.} Mean ratings of the \emph{arg}umentativeness, \emph{cou}ntering, and \emph{rel}evance of the counterarguments generated by four LLMs for 50 test instances from \citet{hua:2018}. The LLMs were evaluated after training either on the training set of the target task, or on the training sets of all other tasks. The best result per criterion is marked bold.}
	\label{table-generation-manual-evaluation}
\end{table}

Table~\ref{table-generation-manual-evaluation} presents the mean ratings, showing that \texttt{Mistral-22b} is judged best, both when trained on the target task and on all other tasks. This corroborates our findings in Section~\ref{sec:leave-one-task-out}. Notably, the performance generally decreases across all dimensions without in-task training: for argumentativeness by 0.807 to 0.126, for countering by 0.686 to 0.040, and for relevance by 0.560 to 0.450. These results underscore the importance of task-specific training for optimal performance on counterargument generation. Furthermore, the higher cross-task performance on the countering dimension compared to argumentativeness suggests that LLMs without task-specific training are slightly better at generating opposing stances than at grounding these stances as well-reasoned arguments.
\section{Follow-up Experiment: Skill Transfer}
\begin{table*}[t]
	\small
	\centering
	\renewcommand{\arraystretch}{0.925}
	\setlength{\tabcolsep}{4pt}
	\begin{tabular}{lcccccc} 
		\toprule
		  \textbf{Training Skill} & \textbf{Mining}    & \textbf{Perspective}  & \textbf{Quality} & \textbf{Reasoning} & \textbf{Generation}& \textbf{Macro}\\ 
		
		\cmidrule(l@{2pt}r@{2pt}){2-2} \cmidrule(l@{2pt}r@{2pt}){3-3} \cmidrule(l@{2pt}r@{2pt}){4-4}  \cmidrule(l@{2pt}r@{2pt}){5-5} \cmidrule(l@{2pt}r@{2pt}){6-6} \cmidrule(l@{2pt}r@{2pt}){7-7}	
				& \textbf{Unit}    & \textbf{Argument}  & \textbf{Argument} & \textbf{Ad-hominem} & \textbf{Counterargument}& \textbf{All}\\ 
				& \textbf{Segmentation}    & \textbf{Similarity}  & \textbf{Rating} & \textbf{Detection} & \textbf{Generation}&\\
		\midrule

								  
								  On the target task	 	  & 	0.270	& 	0.636	& 0.554 	& 0.912		& 0.825		& 0.639\\
								  On all other tasks	 	  & 	0.167	& 	0.468	& 0.529 	& 0.890 	& 0.791		& 0.569\\
								  \midrule
								  Argument Mining	 & 	\textbf{0.194}	& 	0.479		& \textbf{0.552} 	& 0.908 	& \textbf{0.803}		& 0.587\\
								  
								  Perspective Assessment  	  & 	0.183	& 	\textbf{0.598}		& 0.540		& 0.893		& 0.799 	& \textbf{0.603}	\\
								  Argument Quality Assessment  	  	  & 	0.118	& 	0.541	& 0.540 	& 0.904		& 0.801		& 0.580  \\
								 Argument  Reasoning  	 			  & 	0.097	& 	0.486	& 0.534 	& \textbf{0.909}		& 0.800		& 0.565 \\
								  Argument Generation  	 			  & 	0.058	& 	0.524	& 0.544 	& 0.906		& 0.799		& 0.566  \\
		\bottomrule
	\end{tabular}
	\caption{\textbf{Skill transfer results.} Performance of \texttt{Mistral-22b} on the five target tasks, when fine-tuning a LoRa adapter on the training sets all tasks of each skill and then on the training set of the target task. The five target tasks are the same as in Table~\ref{table-results-leave-one-task}.}
	\label{table-skill-transfer-results}
\end{table*}

The results of the leave-one-task-out experiments indicate that task-specific training is required for optimal performance on some tasks, such as counterargument generation \cite{hua:2018}. In this section, we therefore study which training skill most effectively transfers to each of the target tasks. This provides an intuition for how to tailor an optimal training set for a given target task.

\subsection{Skill Transfer Setup}

To explore skill transfer, we fine-tune a LoRa adapter on the training data of each skill, fine-tune it further on the target task, and evaluate it on the test set of the target task. In this way, we can assess which skill prepares LLMs best for unseen tasks. We conduct this experiment using \texttt{Mistral-22b}, since it showed the best generalization capabilities across the evaluated models in the leave-one-task-out experiment. We employ a continual learning setup, where we fine-tune the LoRa adapter for five epochs on the training skill and further fine-tune it on the target task.\footnote{We utilize the hyperparameters for the leave-one-task-out experiment for fine-tuning on a training skill and the in-task experiments for fine-tuning on the target tasks.} For comparison, we compare the results of \texttt{Mistral-22b} in the skill transfer experiment to the results of \texttt{Mistral-22b} in the in-task and leave-one-task-out experiment. 
\subsection{Results}

Table~\ref{table-skill-transfer-results} compares the results of \texttt{Mistral-22b} of the skill transfer experiment to the in-task and leave-one-task-out results. We observe that training on the same skill of the target task achieves the best performance for unit segmentation, argument similarity, and ad-hominem detection, with F$_1$-scores of 0.194, 0.598, and 0.909, respectively. However, on counterargument generation and argument rating, argument mining is a better transfer skill than the target task's skill. This shows the importance of training on argument extraction tasks for counterargument generation and quality assessment tasks.

Since training on the same training skill of a target task is expected to bring good results on the target task, we turn our focus to the best training tasks while excluding the same training skill. For unit segmentation~\cite{habernal:2015}, perspective assessment is the best transfer skill when excluding argument mining, outperforming training on all other tasks (0.183 vs.\ 0.167). This shows that tasks related to understanding the perspective of an argument are effective training tasks for extracting argument units. This is likely because argument units typically feature perspective-taking language, such as stance or frames.

For argument similarity, quality assessment is the most effective transfer skill, exceeding the performance obtained by training on all training tasks (0.468 vs.\ 0.541). This likely stems from the pairwise framing of some quality tasks, which resembles the input format of argument similarity. By examining the best training skill for ad-hominem detection, we notice that argument mining is the most effective transfer skill, with a drop of 0.001 in comparison to training on the same target tasks' skill (argument reasoning). The competitive performance achieved by fine-tuning on argument mining for ad-hominem detection positions argument mining as the best training skill on four of the five target tasks (all but argument similarity).

\section{Conclusion}

The ability of LLMs to understand, analyze, and generate arguments is essential to counter societal harms, to identify fallacious reasoning, and much more. To enable systematic training and evaluation of computational argumentation skills, we introduce ArgBench, a benchmark tailored to LLMs that provides a unified evaluation protocol for 46 existing tasks from five skills: argument mining, perspective assessment, argument quality assessment, argument reasoning, and argument generation. Thereby, it serves as the most comprehensive computational argumentation benchmark available to date. While most tasks have been proposed before the LLM era, we still expect a convincing LLM to tackle them effectively.

In two experimental setups, we have presented empirical insights into the few-shot and chain-of-thought prompting performance of five LLM families on the benchmark and into their generalization ability of computational argumentation skills obtained from fine-tuning. Our prompting results suggest that performance increases with model size, especially for mining, assessing perspectives, and reasoning. In contrast, assessing argument quality is the most challenging skill for LLMs, with chain-of-thought being best for it. Cross-task generalization remains challenging, but seems promising for some quality assessment and reasoning tasks. Manual evaluation of LLMs' output in one generation task corroborates our finding that task-specific fine-tuning is needed for optimal performance in some tasks. Future research on ArgBench may study how to select the best training tasks for a target task.


\section{Limitations}

The main goal of this paper is to provide a comprehensive and unified benchmark for LLM-based approaches to computational argumentation. To reach this goal, we had to make a few compromises that we cover in the following: 

\paragraph{Unified Modeling of Tasks}
Computational argumentation is highly diverse in input and output requirements, partly even notably varying across different datasets for the same task. To provide a common LLM-adjusted interface to all tasks, we needed to slightly reframe certain tasks, as indicated above. This included simplifying argument units to the sentence level, reducing the predefined labels for open tasks such as frame identification~\cite{ajjour:2019} to six labels, or approaching an argument ranking task~\cite{gleize:2019} in a comparative way. We refrained from using specific output formats, such as JSON, to avoid limiting approaches in what they create and not to favor model families optimized toward these formats. 

\paragraph{Coverage of Tasks}
While we present the largest and most comprehensive collection of computational argumentation tasks to our knowledge, some existing datasets have not been included for different reasons. Some have application-specific argument models, such as context-dependent claim detection. \cite{levy:2014}. Others seem rather specific for a general benchmark, such as appropriateness rewriting \cite{ziegenbein:2024}, or their data is not allowed to be redistributed, such as for argument specificity classification \cite{durmus:2019}.

\paragraph{Coverage of LLMs}
The results of the baseline prompting and leave-one-task-out experiments presented above required extensive computation. For space and computational complexity reasons, we could not include model sizes that exceed 32 billion parameters and focused on studying models with different architectures and post-training techniques. To explore their general impact, we included five open-weight model families. While we contrast small and large models in this work, we cannot exclude that other model families or sizes behave differently.

\paragraph{Unified of Evaluation of Outputs}
By nature, LLMs may provide any string as output and cannot be strictly forced to match the output format aimed for. This particularly holds for recent model families, such as \texttt{DeepSeek-R1}, pretrained to carry out chain-of-thought reasoning as part of their output generation. In our evaluation setup, we foster a clear output specification by ending with ``Output:'', as indicated above. Moreover, we used regular expressions to match the output of the different LLMs to the desired format for segmentation tasks (i.e., a list of label and span pairs separated by line breaks). 

\paragraph{Coverage of Measures}

To further support a unified evaluation setting, we opted for as common and as few as possible evaluation measures in our benchmark. Four of the five argumentation skills we cover could be mapped to a setup that allows for macro F$_1$-score, which we see as appealing to obtain comparability even across the respective tasks. As for most families of text generation tasks, a reliable evaluation of argument generation would actually require human judgments, which is not doable in a benchmark. We thus decided to use one of the most common measures of semantic similarity (BertScore). Additionally, we conducted a targeted manual evaluation of LLM outputs for counterargument generation~\cite{hua:2018} to capture human perception of the generated text. We explicitly refrained from LLM judges in the evaluation, as they may be outdated soon and conceptually conflict with the idea of comparing different LLMs to each other. 

\bibliography{task-specific.bib}

\appendix
\section{Appendix}
\label{manual-evaluation-details}
The appendix details the error analysis of the \textit{quality assessment} and \textit{argument reasoning} tasks (\ref{sec:error-analysis}), describes the task selection and normalization for ArgBench (\ref{benchamrk-details}), and provides implementation details of our experiments (\ref{sec:implementation}). Finally, it lists experimental results on more models and provides more information about the manual evaluation of counterargument generation (\ref{sec:extra-experiments}).

\begin{table*}[htb]
    \centering
    \small
    \renewcommand{\arraystretch}{0.95}
    \setlength{\tabcolsep}{1.5pt}
    \begin{tabular}{@{}lccccccccc@{}}
\toprule
\textbf{Task (Dataset)} & \multicolumn{3}{c}{\texttt{Mistral-22b}} & \multicolumn{3}{c}{\texttt{Qwen3-32b}} & \multicolumn{3}{c}{\texttt{Llama3.3-70b}}\\
\cmidrule(lr{2pt}){2-4} \cmidrule(lr{2pt}){5-7} \cmidrule(lr{2pt}){8-10}
& Zero-shot & Few-shot & CoT & Zero-shot & Few-shot & CoT & Zero-shot & Few-shot & CoT \\
\midrule
\multicolumn{7}{@{}l}{\textbf{Argument Quality Assessment}} \\
Argument Ranking (Claim Revisions) 						& .388 		 & .346 		 & .421 		 & .348 		  & .442			& .560		   	    & .570 & .550 & \textbf{.606}\\
Argument Ranking (IBM Evidence Quality)				    & .351 		 & .350 		 & .360 	  	 & .619 		 & .606 		  	& .602			    & .621 & .365 & \textbf{.664}\\
Argument Ranking (IBM Rank 30k) 						& .517			 & .477  		 & .417 		 & .507 		  & .454 			& .515 			& .535 & .513 & \textbf{.546}\\
Argument Rating (Dagstuhl 15512)					    & .343 		 & .255 		 & .360		 & .467 		& \textbf{.502} 	& .381 			& .414 & .433& .360\\
Argument Rating (GAQ) 									& .340 		 & .209 		 & .384 		 & .388 		& .338			& .204 			& .374 & .262& \textbf{.404}\\
Improvement Suggestion (Claim Revisions) 				& .454 		 & .502		 & .440 		 & .448		  & .503	& \textbf{.531} 			&.466 & .508& .499\\
Controversy Scoring (CMV Ad-hominem) 					& .202 		 & .256		 & .313 		 & .327 		 & .357 	& .307 			& .414 & \textbf{.512}& .502\\
Reasonableness Scoring (CMV Ad-hominem) 				& .129 		 & .134		 & .215 		 & .175 		 & \textbf{.282}	& \textbf{.282} 			& .085 & .171& .077\\
Suboptimal Claim Detection (Claim Revisions) 			& .296 		 & .268 		 & .268 		 & .270 		 & .316 	& \textbf{.484} 			& .293 & .451& .292\\
\midrule
Macro-average 													& .335			 & .311 		& .353			 & .394			 & .422			& .430			& .419 & .418 & \textbf{.439}\\
\midrule
\addlinespace
\multicolumn{7}{@{}l}{\textbf{Argument Reasoning}} & & &\\
Ad-hominem Detection (CMV Ad-hominem) 						& .522 		& .815			 & .757 		 & .825 		  & .842	& .643 			& .853 &\textbf{.873} & .858\\
Fallacy Detection (ElecDeb60to20) 						& .027 		& .121 		 & .044 		 & .060 		  & .075 			& \textbf{.126} 	& .086 & .079 &.068\\
Fallacy Detection (Logic) 								& .283 		& .368 		 & .328 		 & .402 		  & .427	& .214 			& .483 & \textbf{.514}& .300\\
Scheme Classification (ArgU) 							& .242 		& .231 		 & .268		& .283		 	 & .253 			& .134				& \textbf{.332} & .257&.203\\
Warrant Identification (SemEval'18) 					& .652 		& .779 		 & .753 		 & .850	 	 & .876			& .673 			& .865 & \textbf{.908}& .549\\
\midrule
Macro-average 													& .345 		& .463 		 & .430		& .484          & .495 			& .358				& .524 & \textbf{.526} & .396\\
\bottomrule
\end{tabular}
    \caption{\textbf{Prompting error analysis.} Macro F$_1$-scores performance of \texttt{Mistral-22b}, \texttt{Llama3.3-70b}, and \texttt{Qwen3-32b} across all argument quality assessment and argument reasoning tasks. The macro-average over the tasks of each skill, which represents the score for the skill, is also reported. The best model and prompting technique per task (row) is highlighted.}
    \label{tab:error-analysis}
\end{table*}

\subsection{Error Analysis}
\label{sec:error-analysis}

The prompting experiments reveal substantial room for improvement across all evaluated models and skills. This is particularly evident in quality assessment and reasoning tasks, even for the highest-performing models. Table \ref{tab:error-analysis} lists the performance of \texttt{Mistral-22b}, \texttt{Llama-3.3-70b}, and \texttt{Qwen3-32b} across all tasks within these two skills. The results show that LLMs are good at warrant identification~\cite{habernal:2018a} and ad-hominem detection~\cite{habernal:2018b}. On the other hand, LLMs struggle with reasonableness scoring~\cite{habernal:2018b} and reasoning tasks such as scheme classification~\cite{saha:2023b} and fallacy detection~\cite{goffredo:2023, jin:2022}. In the following, we describe typical errors exhibited by the three models. 

\begin{table*}
\small

\renewcommand{\arraystretch}{0.7}

\begin{tabular}{p{2.5cm}p{12.5cm}}
	
	\toprule
	\textbf{Instructions} & A fallacy a failure in reasoning which renders an argument invalid. Classify the following sentence into one of the following fallacies: \newline
	[Definitions for fallacies are omitted for sake of space. Look up the repo for details] \newline
	Think step by step and prepend your output with Output: \\
	\toprule
	
	\multicolumn{2}{c}{\textbf{Instance 1 (Correct Prediction)}}\\
	
	\textbf{Input} &Sentence: The meteorologist predicted the wrong amount of rain for May. Obviously the meteorologist is unreliable.\\
	\addlinespace
	\textbf{Ground truth}  & \texttt{faulty generalization}\\
	
	\addlinespace
	\texttt{Qwen3-32b}: & <think>
	Okay, let's see. The sentence is: "The meteorologist predicted the wrong amount of rain for May. Obviously the meteorologist is unreliable." I need to figure out which fallacy this is.\newline
	First, the person is saying that because the meteorologist was wrong once (in May), they're unreliable. That sounds like a faulty generalization. Like, one instance isn't enough to conclude that the meteorologist is always wrong. But wait, maybe there's another angle. Is there an ad-hominem here? Because they're attacking the meteorologist's reliability instead of the argument? Hmm, but the argument is about the prediction. Wait, the person is using the incorrect prediction to say the meteorologist is unreliable. But if they're using a single instance to generalize, that's a faulty generalization. So maybe that's the main issue here. The person is taking one wrong prediction and concluding the whole person is unreliable. That's a classic case of hasty generalization, which is a type of faulty generalization. So I think the answer is Faulty Generalization.\newline
	</think>\newline
	Output: Faulty Generalization\\
	\midrule
\multicolumn{2}{c}{\textbf{Instance 2 (Incorrect Prediction)}}\\

\textbf{Input} & Sentence: I really deserve an “A” on this paper, professor. Not only did I study during my grandmother’s funeral, but I also passed up the heart transplant surgery, even though that was the first matching donor in 3 years.\\
\addlinespace
 \textbf{Ground truth}  &     \texttt{appeal to emotion}
\\
\addlinespace
\texttt{Qwen3-32b}: & <think>
Okay, let's see. The user wants me to classify the given sentence into one of the listed fallacies. The sentence is: "I really deserve an “A” on this paper, professor. Not only did I study during my grandmother’s funeral, but I also passed up the heart transplant surgery, even though that was the first matching donor in 3 years."\newline
First, I need to understand what the speaker is doing here. They're trying to argue that they deserve an A by citing extreme examples of their dedication. Studying during a funeral and refusing a life-saving surgery. But these examples are clearly exaggerated and possibly false. The key here is whether the argument is using irrelevant or extreme examples to support the claim.\newline
Looking at the fallacy options. Ad-hominem is attacking the person, which doesn't fit here. Appeal to emotion might be a possibility because they're trying to evoke sympathy, but the main issue is the use of extreme, possibly false, examples. Faulty Generalization is about drawing conclusions from insufficient samples, but that's not the case here. False Causality is implying a cause-effect without evidence, but the argument isn't about causation.\newline
The Fallacy of Extension is attacking an exaggerated version of the opponent's position, but here the speaker is exaggerating their own actions. Wait, the Fallacy of Extension is when you misrepresent someone's argument by exaggerating it. But in this case, the speaker is exaggerating their
\\	
	\midrule
\multicolumn{2}{c}{\textbf{Instance 3 (Incorrect Prediction)}}\\

\textbf{Input} & Sentence: There must be objective rights and wrongs in the universe. If not, how can you possibly say that torturing babies for fun could ever be right?\\
\addlinespace
\textbf{Ground truth}  &     \texttt{appeal to emotion}
\\
\addlinespace
\texttt{Llama-3.3-70b}: & The given sentence can be classified as an example of a false dilemma. The argument presents only two options: either there are objective rights and wrongs in the universe, or torturing babies for fun could be considered right. this is a false dichotomy because it ignores the possibility of other moral frameworks, such as moral relativism or subjective morality, where the morality of an action is determined by cultural, personal, or situational factors, rather than by objective universal standards. the sentence implies that if one does not accept the existence of objective rights and wrongs, they must condone or consider acceptable the torture of babies for fun, which is not a necessary or logical consequence. this oversimplification of complex moral issues into only two extremes is characteristic of a false dilemma. \\
\midrule
\multicolumn{2}{c}{\textbf{Instance 4 (Incorrect Prediction)}}\\

\textbf{Input} & Sentence: Thousands of pets are left homeless and hungry because lousy people turn them out on the streets instead of taking care of them.\\
\addlinespace
\textbf{Ground truth}  &     \texttt{appeal to emotion}
\\
\addlinespace
\texttt{Mistral-22b}: & Output: Fallacy of Relevance: The sentence introduces a conclusion about pets being homeless and hungry, but it does not address the actual reasons behind this issue. It simply blames "lousy people" without providing evidence or a logical connection between the two. Additionally, it does not consider other possible factors that could contribute to the problem. \\

	\bottomrule
\end{tabular}
\caption{Examples of correct and incorrect predictions for \texttt{Qwen3-32b}, \texttt{Mistral-22b}, and \texttt{Llama-3.3-70b} with Chain-of-thought in the task Fallacy Detection~\cite{jin:2022}.}
\label{table-fallacy-detection-error-analysis}
\end{table*}

Analysis of model outputs reveals that \texttt{Mistral-22b} exhibits a label bias toward the positive or high-magnitude labels in almost all tasks. Examples of these positive labels are \texttt{very controversial} for controversy scoring or \texttt{quite reasonable} for reasonableness scoring. The problem is especially pronounced in the three pairwise argument ranking tasks where \texttt{Mistral-22b} with zero-shot prompting selects \texttt{better} over \texttt{worse} in 2,669 out of 3,000 (89\%) cases. This bias is further amplified in few-shot prompting where \texttt{Mistral-22b} predicts \texttt{better} in 2,801 out of 3,000 (93\%) cases. \texttt{Llama3.3-70b} predicts \texttt{better} in 1,471 (49\%) cases in zero-shot-prompting and in 2,294 (76\%) cases in few-shot-prompting, indicating a slightly lower bias than \texttt{Mistral-22b} towards positive labels as well.  

In argument rating tasks \cite{ng:2020,wachsmuth:2017b}, \texttt{Mistral-22b} tends toward the middle labels: \texttt{average} or \texttt{medium}. In argument rating~\cite{wachsmuth:2017b}, the models have to classify an argument on a quality dimension (e.g., Rhetoric) into either \texttt{low}, \texttt{average}, or \texttt{high}, indicating how good the argument is on a given dimension. \texttt{Mistral-22b} responds with the middle label in 1,666 out of 2,000 (83\%) cases in zero-shot prompting in the two argument rating tasks. In few-shot prompting, \texttt{Mistral-22b} predicts \texttt{medium} in 1,135 out of 2,000 (57\%) cases. The tendency of the model to select the middle label suggests the inclination of the model to pick the ``safe'' option. This bias may stem from the inherent subjectivity of argument rating, where argument quality often depends on the target audience.

We identify three primary error sources in chain-of-thought prompting: incomplete reasoning chains, difficulties in processing emotional language, and failures in instruction following. 

Incomplete reasoning occurs when the model exhausts all the token budget without generating the final output. Specifically, \texttt{Qwen3-32b} entered unfinished reasoning loops in 5,845 out of 7,720 (76\%) argument quality assessment instances and in 2,299 out of 2,953 (78\%) argument reasoning instances.\footnote{Notice that in these cases we assigned a random label as the output of the model. We did not optimize the max output tokens for each model separately to have a fair comparison between the models.} Table~\ref{table-fallacy-detection-error-analysis} provides exemplary predictions for \texttt{Qwen3-32b}, \texttt{Mistral-22b}, and \texttt{Llama-3.3-70b} in fallacy detection~\cite{jin:2022}. Instance 2 illustrates a typical error where \texttt{Qwen3-32b} iterates over multiple candidate fallacies, hesitates to assign the correct \texttt{appeal to emotion} label, and eventually stops while reflecting on an irrelevant candidate (\texttt{fallacy of extension}). This behavior may stem from excessive self-reflection, where \texttt{Qwen3-32b} exhausts the token budget by weighing multiple labels despite having initially identified the correct one. Additionally, the lengthy instructions required for fallacy detection and scheme classification~\cite{jin:2022, saha:2023b, goffredo:2023} may exacerbate this token exhaustion.

Handling emotional language is another source of errors that LLMs struggle with in argument reasoning tasks. Instances 3 and 4 show arguments that exemplify an \texttt{appeal to emotion}, but \texttt{Llama-3.3-70b} and \texttt{Mistral-22b} incorrectly categorize them as \texttt{false dilemma} and \texttt{fallacy of relevance}, respectively. In Instance 3, \texttt{Llama-3.3-70b} focuses on the comparative structure of the argument that presents two extreme options, thereby overlooking the appeal to audience sympathy. Instance 4 shows a similar example where the model focuses on the logical connection between the conclusion and the reason and ignores the emotional content of the whole argument.

Finally, we observed frequent instruction following failures in \texttt{Llama-3.3-70b} and \texttt{Mistral-22b}. \texttt{Llama-3.3-70b} ignores the formatting instructions in some cases, exclusively in chain-of-thought prompting. An example of such instances is Instance 3. In this example, \texttt{Llama-3.3-70b} does not use the prefix ``Output:'' and does not generate the fallacy at the end of the output. Other failures include using labels that are not listed in the task instructions or inventing a new label format (e.g., ``**Output: ...**''). Appendix \ref{sec:implementation} provides more details on output extraction.

\subsection{Task Selection and Formulation}
\label{benchamrk-details}

In the following, we outline all tasks in the ArgBench benchmark for each of the five skills that we derived from the datasets listed in Table~\ref{table-datasets}. While not critical to the paper itself, we detail the selection and formulation of the tasks to provide insights into the benchmark creation process. Table~\ref{table-prompting-set-statistics} contains the size of the test sets in ArgBench for each skill. Further below, Tables~\ref{table-prompts} to~\ref{table-prompts-last} contain the task definitions as well as the input template for an exemplary set of tasks. The full list of task prompts can be found in the git repository.

\begin{table*}[!t]
	\small
	\renewcommand{\arraystretch}{0.91}
	\setlength{\tabcolsep}{2pt}
	\begin{tabular}{@{}lll@{\!\!\!}r@{\,}lll@{}}
		\toprule
		\bf Skill & \bf Task & \bf Dataset  & \multicolumn{2}{c}{\bf Size}& {\bf Genre} & \bf Source \\
		
		\midrule		
		\textbf{Argument}	& Conclusion Extraction			& ECHR					& 42& &Case Law&     \citet{poudyal:2020}\\
		\textbf{Mining}		& Premise Extraction			& ECHR					& 42& &Case Law&     \citet{poudyal:2020}\\
							& Premise Extraction			& IBM-Evidence-Sentence	& 29,429& &Encyclopedia&     \citet{eindor:2020}\\
							& Relation Detection  			& ECHR 					& 42	& &Case Law& \citet{poudyal:2020}\\
							& Relation Identification  		& Microtexts-1 			& 112	& & Student Essays& \citet{peldszus:2014}\\ 	
							& Relation Identification  		& Essays 				& 402	& &Student Essays& \citet{stab:2017a}\\
							& Relation Identification  		& Microtexts-2 			& 171	& &Student Essays& \citet{skeppstedt:2018}\\
							& Relation Identification  		& Political-debates 	& 1,462	& &Spoken Debates& \citet{menini:2018}\\
							& Relation Identification  		& E-rulemaking 			& 731	& &Web Forum& \citet{park:2018}\\
							& Unit Classification  			& Essays 				& 402	& &Student Essays& \citet{stab:2017a}\\
							& Unit Classification  			& Editorials 			& 731	& &News& \citet{alkhatib:2017}\\
							& Unit Classification  			& E-rulemaking 			& 731	& &Web Forum& \citet{park:2018}\\
							& Unit Segmentation   			& WebDiscourse  		& 340	& &Web Forum &{\citet{habernal:2015}}\\
							& Unit Segmentation   			& Editorials  			& 300	& &News& {\citet{alkhatib:2016}} \\
							& Unit Segmentation   			& Essays  				& 402	& &Student Essays& \citet{stab:2017a}\\
							& Unit Segmentation   			& ECHR	 				& 42	& &Case Law& \citet{poudyal:2020}\\


		\midrule
		\textbf{Argument}	& Argument Similarity  			& UKP-Aspect 			& 3,595	& &Mixed& \citet{reimers:2019} \\
		\textbf{Perspective}& Aspect Detection				& UKP-Corpus			& 5,032	& &Mixed& \citet{schiller:2021}\\
		\textbf{Assessment}	& Frame Identification 			& Webis-argument-framing& 12,326& &Online Debates& \citet{ajjour:2019}\\
							& Key Point Matching  			& ArgKP 				& 24,093& &Encyclopedia& \citet{barhaim:2020}\\
							& Stance Classification  		& IBMSC 				& 2,394	& &Encyclopedia& \citet{barhaim:2017}\\
							& Stance Classification  		& Ukp-sentential 		& 25,492& &Mixed& \citet{stab:2018}\\
										
		\midrule
		\textbf{Argument}	& Argument Ranking 				& IBM-Evicence-Quality  & 5,697	 & &Encyclopedia& \citet{gleize:2019}\\
		\textbf{Quality}	& Argument Ranking 				& IBM-RANK-30k  		& 30,497 & &Encyclopedia& \citet{gretz:2020}\\
		\textbf{Assessment}	& Argument Ranking 				& Claim Revisions  		& 587,881& &Online Debates&\citet{skitalinskaya:2021}\\
							& Argument Rating  				& Dagstuhl-15512		& 320	 & &Online Debates& \citet{wachsmuth:2017b}\\ 
							& Argument Rating 				& GAQ			  		& 5,285	 & &Web Forum&\citet{ng:2020}\\
							& Controversy Scoring			& ChangeMyView			&7,242   & &Web Forum& \citet{habernal:2018b}\\
							& Reasonableness Scoring		& ChangeMyView			&7,242   & &Web Forum & \citet{habernal:2018b}\\
							& Suboptimal Claim Detection	& Claim Revisions 		& 198,089& &Online Debates& \citeauthor{skitalinskaya:2023a}\\						
							& Claim Improve.\ Suggestion	& Claim Revisions 		& 198,089& &Online Debates& \citeauthor{skitalinskaya:2023a}\\						
		\midrule	
		\textbf{Argument}	& Ad-hominem Detection 			& Ad-hominem-CMV 		& 1800	 & &Web Forum& \citet{habernal:2018b}\\
		\textbf{Reasoning}	& Fallacy Detection 			& Logic					& 3,800	 & &Mixed& \citet{jin:2022}\\
							& Fallacy Detection 			& ElecDeb60to20			& 1,640	 & &Spoken Debates& \citet{goffredo:2023}\\
							& Scheme Classification 		& ArgU					& 69,427 & &Online Debates& \citet{saha:2023b}\\
							& Warrant Identification 		& SemEval-2018-task12  	&  1,970 & &News & \citet{habernal:2018a}\\

		\midrule 
		\textbf{Argument}	& Argument Generation 			& ArgU 					& 2,990	  & &Online Debates& \citet{saha:2023b}\\
		\textbf{Generation}	& Argument Summarization        & OpenDebateEvidence 	&4,957,726& &Online Debates&\citet{roush:2024}\\
							& Claim Generation		      	& Belief Arguments 		& 51,470  & &Online Debates& \citet{alshomary:2021}\\
							& Claim Optimization			& Claim Revisions 		& 198,089 & &Online Debates& \citet{skitalinskaya:2023b}\\

							& Counterarg.\ Generation 	& ChangeMyView 			& 268,881 &  &Web Forum& \citet{hua:2018}\\
							& Counterarg.\ Generation	& Candela 				&287,152  &  &Web Forum& \citet{hua:2019}\\
							& Warrant Generation			& ART 					& 72,846&    &Narratives& \citet{bhagavatula:2020}\\
							& Warrant Generation			& ARC					& 1,654	&	 &News& \citet{habernal:2018a}\\
							& Warrant Generation			& Ideological Debates 	& 494	&    &Online Debates& \citet{boltuzic:2016}\\
							& Warrant Generation			& Micortext-1			& 112	&	 &Student Essays& \citet{becker:2020}\\	
		\bottomrule
	\end{tabular}
	\caption{The list of all computational argumentation datasets in ArgBench together with the numbers of input units.}
	\label{table-datasets}
\end{table*}

\paragraph{Argument Mining}

Argument mining is the process of extracting arguments from natural language text. Traditional mining pipelines start with \emph{unit segmentation}, where the task is to extract argument units (also callsed argument \emph{components}) from an input document, that is, text spans that have an argumentative function. In our benchmark, we include four widely-used datasets for the task \cite{habernal:2015,alkhatib:2016,stab:2017a, poudyal:2020} and define the instruction for them as follows: 

\texttt{Split the documents into spans that are argumentative and those that are not-argumentative, separated with a line break.} 


After segmenting a document into units, \emph{unit classification} is the task to assign one of multiple labels to each unit that describes the role of a unit in an argument  (e.g., \texttt{major claim}, \texttt{claim}, or \texttt{premise}) or the type of evidence (e.g., \texttt{anecdote}, \texttt{expert opinion}, or similar).%
\footnote{In our benchmark, we assume each argument unit to span one sentence to allow an easy and consistent evaluation.} 
To capture task diversity, we include three known datasets with texts of different genres \cite{stab:2017a,alkhatib:2017,park:2018} and different label schemes (see also Table~\ref{table-modeled-tasks} in the main paper). We prepend the argument unit with a context window to classify the argument unit. The context window spans five or more sentences that contain the input argument unit. We chose the maximum number of sentences as a context window size, while maintaining that the input's length for an instance does not exceed 1024 tokens. 
Simpler variants of argument unit classification split the task into several binary classification tasks for the target unit types. ArgBench therefore also contains three of these tasks: \emph{conclusion extraction} in case law~\cite{poudyal:2020}, \emph{premise extraction} in case law~\cite{poudyal:2020}, and premise extraction in Wikipedia~\cite{eindor:2020}. We add context of at least five sentences that contain the input sentence to these as well. 

The third mining step is \emph{relation detection}, which means to decider whether there is an argumentative relation between two argument units that makes them part of the same argument. In isolation, we include in case law as defined by \citet{poudyal:2020}, where the input is a sentence pair and the output is either \texttt{related} or \texttt{not-related}. Negative sentence pairs (\texttt{not-related}) are all possible sentence pairs within a window of five sentences. We append the input sentence pair with a context of five instances that contain the candidate sentence pairs as input for this task.

After detecting the relation between argument units, the next step is \emph{relation identification} \cite{peldszus:2014}, where the task is to classify the relation between two argument units into types that are defined in the argument model. Aiming for diversity again, ArgBench includes four classical relation identification datasets \cite{peldszus:2014,stab:2017a,skeppstedt:2018, park:2018}. The input is a pair of argument units accompanied by a context window, and the output is (\texttt{reason} or \texttt{evidence}) in E-rulemaking~\cite{park:2018} and (\texttt{support} or \texttt{attack}) in the other three datasets. As a context window, we take all sentences between the candidate sentence pair. A similar task is to identify the relation between two arguments instead of argument units. In ArgBench, we include the political debates dataset for this task \cite{menini:2018}, which takes as input an argument pair and classifies their relation into \texttt{support}, \texttt{attack}, or \texttt{not-related}.

\paragraph{Argument Perspective Assessment}

This skill covers the identification of all concepts describing the perspective of an argument on a given topic. As this comes in many task shades, we aim to reflect this variety in Benchmark, rather than having many datasets for each task.

One main perspective assessment task is \emph{stance classification}, which means to determine the position of an argumentative text toward a topic (e.g., \texttt{pro}, \texttt{con}, or \texttt{neutral}). As two known but very different examples, we includes essays \citep{stab:2017a} and the Wikipedia sentences from IBMSC~\cite{barhaim:2017}. The input for stance classification tasks in ArgBench is an argument and a topic, and the output is the stance.

In \emph{frame identification}, the goal is to extract the aspect of a topic that a given argument discusses, which we represent with one widely used dataset~\cite{ajjour:2019}. For a unified frame identification setup, we provide an argument and a frame and ask the model to output \texttt{match} or \texttt{no-match}. We sample five wrong frames on the same topic for each argument in addition to the right frame associated with each argument. In case the count of frames for a topic is less than five, we sample frames from other topics. The rationale behind choosing five negative frames is that on average a topic is framed using six frames in the Webis-argument-framing dataset. The related task of \emph{aspect detection}, represented by the data of \citet{schiller:2021}, is defined by asking the model to segment an input document into spans that correspond to a specific aspect or not. Similar to unit segmentation, this task is evaluated by measuring precision, recall, and F$_1$-score of exact matches of the spans labeled with aspect with the ground truth spans labeled with aspect.

A more specific task is to assess \emph{argument similarity}. In case of the data of \citet{reimers:2019}, the input is a pair of arguments and the output one of four labels: \texttt{high similarity}, \texttt{some similarity}, \texttt{no similarity}, and \texttt{different topics}. Finally, \emph{key point matching} means to decide whether an argument aligns with a key point or not and has been introduced in  dataset from a successful shared task \cite{barhaim:2020}. It takes as input an argument and a key point and returns \texttt{match} or \texttt{no-match}. 

\begin{table}[t]
	\small
	\renewcommand{\arraystretch}{1.0}
	\setlength{\tabcolsep}{5pt}
	\centering
	\begin{tabular}{ll}
		\toprule
		\textbf{Skill}& \textbf{Instances}\\
		\midrule
		Argument Mining & 10,039 \\
		Argument Perspective Assessment & 5,758 \\
		Argument Quality Assessment & 7,720 \\
		Argument Reasoning & 2,953 \\
		Argument Generation & 7,325 \\
		
		\bottomrule
	\end{tabular}
	\caption{Count of instances per skill in the prompting experiment. }
	\label{table-prompting-set-statistics}
\end{table}

\paragraph{Argument Quality Assessment}

The quality of an argument is crucial to persuade an audience of the perspective that the argument holds. This task can be modeled as an \emph{argument rating} task, that is, to assign a value from a predefined scale (e.g., \texttt{low}, \texttt{medium}, or \texttt{high}) for a given quality dimension (e.g., reasonableness). For two of the most known respective datasets \cite{wachsmuth:2017b,ng:2020}, we provide as input the argument, the quality dimension (e.g., effectiveness), and the definition of the quality dimension as provided in the paper. In addition, we analogously define two dimension-specific tasks in ArgBench, which are studied on ChangeMyView~\cite{habernal:2018b} (\emph{reasonableness scoring} and \emph{controversy scoring}).

In contrast to absolute ratings, argument quality can be framed in a relative manner, where an LLM judge which of argument between an argument pair is better. We also represent these \emph{argument ranking} tasks with two datasets \cite{gleize:2019,skitalinskaya:2021}, providing the two pieces of evidence or arguments as input. We ask the model to output whether the first argument is \texttt{better} or \texttt{worse} than the second argument. Assessing the quality of arguments is also modeled as a scoring task, where the input is a list of arguments and the output is a list of scores that reflects their relative quality~\cite{gretz:2020}. Defining the task as scoring a long list of arguments violates the design constraint of limiting the the input instance to 1024 tokens. To ensure a computationally feasible setup within ArgBench, we integrate this task in ArgBench by modeling it as a pair-wise argument ranking task. For each argument in the dataset, we sample an argument on the same topic and consider it \texttt{better} if it has a higher score, and \texttt{worse} otherwise.

With their data, \citet{skitalinskaya:2023a} have proposed two specific tasks that we further include. One is \emph{suboptimal claim detection} (i.e., detecting weak claims in need of revision), which is modeled as a binary classification task. The input for this task is the claim, and the output is either \texttt{improvable} or \texttt{not-improvable}. The other is \emph{claim improvement suggestion}, which is to make one of multiple revision suggestions for a claim. We define the task as a binary classification task, where the input is the claim and one of the four revision suggestions, and the output is \texttt{match} or \texttt{no-match}. The four revision suggestions that are included in this work are: clarification, typo/grammar, links, and other. 

\begin{table*}
\small
\renewcommand{\arraystretch}{0.825}
\begin{tabular}[t]{p{2cm}m{9cm}m{3.5cm}}

\toprule
\textbf{Task} & \textbf{Task Definition} & \textbf{Authors}\\
\midrule
Conclusion \newline Extraction & Given the following document, Judge if the following sentence is a conclusion or not. A conclusion is a controversial statement and the central component of an argument. Answer only with conclusion and No-conclusion.  & \cite{poudyal:2020} \\

\\
&  \texttt{Instance Format}  &\\
 &  Sentence:  &\\
 &  Document:  &\\
\midrule
Premise \newline Extraction & Given the following document, Judge if the following sentence is a premise or not. A Premise is a reason for justifying or refuting a claim. Answer only with Premise or No-premise.& \cite{poudyal:2020} \\
\\
&  \texttt{Instance Format}  &\\
&  Sentence:  &\\
&  Document:  &\\

\midrule
Premise \newline Extraction & Judge if evidence can be used to support or attack the motion. Possible outputs: Accept if evidence can be an argument to support or attack the motion or Reject if the evidence can not be used to attack or support the motion. Only output Accept or Reject.& \cite{eindor:2020}\\
\\
&  \texttt{Instance Format}  &\\
&  Motion:  &\\
&  Evidence:  &\\

\midrule
Relation \newline Detection & Given the following document and two sentences, your task is to judge whether they are part of the same argument. An argument consists of a conclusion and multiple premises. Your task is to judge whether the two sentences are part of the same argument, where one sentence supports or attacks the other. Output related if there is an argumentative relation between the two sentences or not-related if not. Only output related or not-related. & \cite{poudyal:2020} \\
\\
&  \texttt{Instance Format}  &\\
&  Sentence 1:  &\\
&  Sentence 2:  &\\
&  Document:  &\\
\midrule
Relation \newline Identification & Given the following essay and the appended source and target argument units that appear in the essay. Output Support if the source argument unit supports the target argument unit, or output Attack if the source attacks the target. Only output Support or Attack. & \cite{peldszus:2015} \\
\\
&  \texttt{Instance Format}  &\\
&  Source:  &\\
&  Target:  &\\
&  Document:  &\\
\midrule
Unit \newline Classification & Given the following document and span, classify the span that appears in the document into the following argument unit types: Common Ground, Assumption, Testimony, Statistics, Anecdote, or Other. Common Ground: is common knowledge, a self-evident fact, an accepted truth, or similar. [Definitions for unit types are omitted for sake of space. Look up the github repo for more details.] & \cite{alkhatib:2016} \\

\\
&  \texttt{Instance Format}  &\\
&  Span:  &\\
&  Document:  &\\

\midrule
\end{tabular}

\caption{Prompt definition and instance format for each task in our benchmark. Very similar tasks are dropped and only the task with the oldest paper is kept. Remaining tasks are in the next three tables.}
\label{table-prompts}

\end{table*}
\begin{table*}
\small
\renewcommand{\arraystretch}{0.825}
\begin{tabular}[t]{p{2cm}m{9cm}m{3.5cm}}

\toprule
\textbf{Task} & \textbf{Task Definition} & \textbf{Authors}\\
\midrule

Unit \newline Segmentation & Given the following document, split all of the document into argumentative units and non-argumentative units. An argumentative unit is a statement that has an argumentative function for example a claim or anecdote. An argumentative unit may span a clause, a complete sentence, multiple sentences, or something in between. Prepend each argumentative unit with argumentative: and spans that are not Argumentative with Non-argumentative:. Output the extracted spans as they are ordered in the given document and separate them by a new line. Do not add a new formatting or enumeration also do not rephrase the argument units. Order the output spans as they appear in the document. & \cite{habernal:2015} \\
\\
&  \texttt{Instance Format}  &\\
&  Document:  &\\
\midrule

Argument \newline Similarity & Judge if the argument pairs are similar (on the same topic and cover similar aspects) or dissimilar (e.g., cover different topics or different aspects). Possible outputs: High similarity if two arguments are very similar, Some Similarity if two arguments are somewhat similar, No Similarity if two arguments are not similar, or Different Topics if two arguments belong to different topics. Only output High Similarity, Some Similarity, No Similarity, Different Topics. & \cite{reimers:2019} \\

\\
&  \texttt{Instance Format}  &\\
&  Argument 1:  &\\
&  Argument 2:  &\\

\midrule
Aspect \newline Detection & Given the following argument, split the argument into spans of text that cover an aspect or not. An aspect is a span of the argument that characterizes the argument. Multiple aspects can be found in an argument. Prepend the aspect span with Aspect and the not-aspect span with Not-aspect. Do not rephrase the spans or modify it. Always process the whole argument. Multiple aspects can be found in an argument. In case there is no aspect, simply output the argument with Not-aspect before it. & \cite{schiller:2021} \\

\\
&  \texttt{Instance Format}  &\\
&  Argument:  &\\

\midrule
Frame \newline Identification & Judge if the given frame captures the most salient aspect of the given argument on the given topic. The frame is the main highlighted aspect of the topic which resonate with as specific audience. Possible responses: Match if the argument emphasizes the given frame and No-match if the argument is not emphasized by the frame. Only output Match or No-match. &  \cite{ajjour:2019} \\

\\
&  \texttt{Instance Format}  &\\
&  Topic:  &\\
&  Argument:  &\\
&  Frame:  &\\
\midrule
Key Point \newline Matching & Judge if the following key point summarizes the given argument. A key point is a short talking point. Key points may be viewed as high-level arguments. They should be general enough to match a significant portion of the arguments, yet informative enough to make a useful summary. Possible responses: Match if argument is summarized by key point and No-match if argument is not summarized by key point. Only output Match or No-match. & \cite{barhaim:2020} \\

\\
&  \texttt{Instance Format}  &\\
&  Argument:  &\\
&  Key Point:  &\\
\midrule
Stance \newline Classification & Classify the stance of the following claim into Pro or Con. Answer with Pro if the following claim supports the following topic. Answer with Con if the claim attacks the topic. Only answer with Pro or Con. & \cite{barhaim:2017} \\

\\
&  \texttt{Instance Format}  &\\
&  Topic:  &\\
&  Claim:  &\\

\midrule

\end{tabular}

\caption{Continuation of Table~\ref{table-prompts}. Details are described there.}
\end{table*}
\begin{table*}
\small
\renewcommand{\arraystretch}{0.825}
\begin{tabular}[t]{p{2cm}m{9cm}m{3.5cm}}
\toprule
\textbf{Task} & \textbf{Task Definition} & \textbf{Authors}\\
\midrule

Argument \newline ranking & Given the following argument pairs, is the first argument better or worse than the second argument in terms of quality. Only respond with Better or Worse. & \cite{gretz:2020}\\

\\
&  \texttt{Instance Format}  &\\
&  Argument 1:  &\\
&  Argument 2:  &\\

\midrule

Argument \newline rating & Judge the quality of the argument according to quality aspect. Possible outputs: Low if arguments aspect quality is low, Average if argument's aspect quality is average, High if arguments aspect quality is high. Only output Low, Average, or High. & \cite{wachsmuth:2017b} \\

\\
&  \texttt{Instance Format}  &\\
&  Argument:  &\\
&  Quality Aspect:  &\\
& Quality Aspect Definition: &\\
\midrule
Controversy \newline Scoring & Classify the following post according to its controversy into either Not Really Controversial, Somehow Controversial, or Very Controversial. & \cite{habernal:2018b}\\

\\
&  \texttt{Instance Format}  &\\
&  Post:  &\\

\midrule
Reasonableness \newline Scoring & Classify the following post according to its reasonableness into : Quite Stupid, Neutral, or Quite Reasonable. & \cite{habernal:2018b} \\

\\
&  \texttt{Instance Format}  &\\
&  Post:  &\\

\midrule
Suboptimal \newline Claim \newline Detection & Judge if the following claim can be improved by revising it. Possible outputs: Improvable if revision should be made, Non-Improvable if no revision is necessary. Only output Improvable or Non-Improvable.& \cite{skitalinskaya:2023a} \\

\\
&  \texttt{Instance Format}  &\\
&  Claim:  &\\

\midrule

Claim \newline Improvement \newline Suggestion & Given an argumentative claim, does the following quality issue match the following claim. Available quality issues are Clarification, Typo/Grammar, Links, or Other. If the quality issue matches the claim, output Match. If the quality issue does not apply to the claim, output No-match. Only output Match or No-match. & \cite{skitalinskaya:2023a} \\

\\
&  \texttt{Instance Format}  &\\
&  Claim:  &\\
& Quality Issue: &\\
\midrule

Fallacy \newline Detection & Given the following argument, split the argument into spans that contains one of the following fallacies. A fallacy is a failure in reasoning which renders an argument invalid. In case a span does not contain a fallacy, simply prepend it with No-fallacy. The split spans should be separated by newlines and be output in the exact order they appear in the argument. Add before each span that covers a fallacy the name of the fallacy and a colon. Do not rephrase anything in the argument. Here are the candidate fallacies: [Definitions for fallacies are omitted for sake of space. Look up the repo for details]& \\

\\
&  \texttt{Instance Format}  &\cite{goffredo:2023}\\
&  Argument:  &  \\

\midrule
Fallacy \newline Detection & A fallacy is a failure in reasoning which renders an argument invalid. Classify the following sentence into one of the following fallacies. [Definitions for fallacies are omitted for sake of space. Look up the repo for details] & \cite{jin:2022} \\

\\
&  \texttt{Instance Format}  &\\
&  Sentence:  &\\

\midrule
\end{tabular}
\caption{Continuation of Table~\ref{table-prompts}. Details are described there.}
\end{table*}
\begin{table*}
\small
\renewcommand{\arraystretch}{0.825}
\begin{tabular}[t]{p{2cm}m{9cm}m{3.5cm}}
\toprule
\textbf{Task} & \textbf{Task Definition} & \textbf{Authors}\\
\midrule
Scheme \newline Classification & Classify the following argument according to the following Walton's argument schemes. [Definitions for schemes are omitted for sake of space. Look up the repo for details] Only output one of the following argument schemes: means for goal, goal from means, from consequence, source knowledge, source authority, rule or principle, and other.& \cite{saha:2023b}\\

\\
&  \texttt{Instance Format}  &\\
&  Argument:  &\\

\midrule

Warrant \newline Identification & Given the following reason and claim along with the debate title and a short description of the debate they occur in, identify the correct warrant from two candidates. Warrant 1 and Warrant 2. The warrant explains why the claim follows from the reason. Only output Warrant 1 or Warrant 2.& \cite{habernal:2018b} \\

\\
&  \texttt{Instance Format}  &\\
&  Reason:  &\\
&  Claim:  &\\
\midrule
Argument \newline Generation & Given the following argument type, topic, stance, and facts, generate an argument that holds that stance on the topic and is based on the facts. Facts are real-world concepts, propositions, and knowledge and do not refer to only knowledge-based facts.& \cite{saha:2023b}\\

\\
&  \texttt{Instance Format}  &\\
&  Argument Type:  &\\
&  Topic:  &\\
&  Stance:  &\\

\midrule

Argument \newline Summarization & Provide an abstractive summary/card-tag of the argument made in the document below.& \cite{roush:2024} \\

&  \texttt{Instance Format}  &\\
&  Document:  &\\

\midrule

Claim \newline Optimization & Given the following input argumentative claim with context information on the debate, rewrite the claim such that the output claim improves upon input claim in terms of text quality and argument quality, and preserves the meaning of the claim as far as possible. & \cite{skitalinskaya:2023b} \\
\\
&  \texttt{Instance Format}  &\\
&  Thesis:  &\\
&  Claim:  &\\
\midrule

Conclusion \newline Generation & Given a discussion topic and a collection of topic stances that describe users stance on various issues, generate a claim that is based on the user stance. A claim is a controversial statement and the central component of an argument.& \cite{alshomary:2021}\\

\\
&  \texttt{Instance Format}  &\\
&  Topic:  &\\
&  User Stances:  &\\

\midrule
Counter\newline  Argument\newline  Generation & Write a counterargument to the following original post and take into account retrieved passages related to the post. & \cite{hua:2019} \\

&  \texttt{Instance Format}  &\\
&  Post:  &\\
&  Retrieved Passages:  &\\

\midrule
Counter\newline  Argument\newline  Generation & Given a statement and relevant evidence, generate a counterargument that attacks to the original argument and highlights the given key phrases. & \cite{hua:2018} \\

&  \texttt{Instance Format}  &\\
&  Statement:  &\\
&  Key Phrases:  &\\
&  Evidence:  &\\
\midrule

Warrant\newline  Generation & Given a premise and a claim, generate an enthymem. An enthymem is a reason with which the claim follows logically form the premise. & \cite{chakrabarty:2021} \\

&  \texttt{Instance Format}  &\\
&  Premise:  &\\
&  Claim:  &\\

\midrule

\end{tabular}
\caption{Continuation of Table~\ref{table-prompts}. Details are described there.}
\label{table-prompts-last}
\end{table*}

\paragraph{Argument Reasoning}

For understanding an argument's reasoning, reasoning patterns are often modeled via argumentation \emph{scheme classification}; we include one dataset here \cite{saha:2023a} with six scheme classes (e.g., \texttt{argument from consequences}). \citet{habernal:2018a} introduced the \emph{warrant identification} task for an argument (i.e., why the conclusion follows from the premise), given a valid and an invalid warrant as input. Another line of research is concerned with various types of \emph{fallacy detection}. Aiming to well-represent the spectrum, we include both the data of \citet{habernal:2018b}, where the goal is to detect ad-hominem arguments in the ChangeMyView forum (i.e., attacks on authors of arguments), the one of \citet{jin:2022} who classify fallacies into 13 types (e.g., \texttt{false cause}). We define all these tasks in the benchmark as classification tasks: Given an argument, output its reasoning pattern or the valid warrant as a label.

In contrast, \citet{goffredo:2023} approaches the task by asking the model to segment the input text into six fallacy types. Accordingly, we ask the model to split the text into spans of text that correspond to one of the fallacies and to append the span with the fallacy label. In case a span does not cover a fallacy, the model is required to output \texttt{no fallacy}. For evaluation, we measure exact matches with the ground-truth spans for the six fallacy types and return their macro average.

\paragraph{Argument Generation}

The last skill is to craft an argument or argument unit for some input (e.g., for a statement). Again, we seek to capture the variety of tasks in our ArgBench benchmark.

For \emph{argument generation}, \citet{saha:2023b} first generate a template that is then filled with facts. \citet{hua:2018} and \citet{hua:2019} tackle the \emph{counterargument argumentation} for a given topic. Researchers also approach generating arguments by generating one of its components (either a conclusion or its warrants). Among these, \citet{alshomary:2021} proposed belief-based \emph{claim generation} to obtain claims that have a pro or con position on a big issue. With four different datasets, \citet{chakrabarty:2021} studied the \emph{warrant generation task}, producing an argument's warrant, given its premise and conclusion. \citet{skitalinskaya:2023b} introduced \emph{claim optimization} as the task of revising claims to improve their quality, whereas \citet{roush:2024} aimed for \emph{argument summarization}. Since generation matches the general idea of LLMs, we simply model all taks as completion tasks, given the respective input.

\subsection{Implementation Details}
\label{sec:implementation}

\paragraph{Output Extraction}
To extract the output of the LLMs, we used regular expressions to capture the output of the model and to remove the chain-of-thought reasoning steps. As a first cleaning step, we removed all reasoning clues that are surrounded by ``<think>'' and ``<think/>''. For classification tasks, we first inspect the first or the last tokens of the completed text. If none of the labels are found at the beginning of the generated text, we inspect the text after one of the following indicators: Output:'' or ``<think/>''. In case we could not extract the label, we sample a label from the candidate labels randomly and consider it to be the predicted label. For segmentation tasks, we extract all spans of texts that are prepended with one of the span labels (e.g., \texttt{argumentative} and \texttt{Non-argumentative} for unit segmentation). For generation tasks, we take all the generated text as the output of the model after removing the reasoning steps. In case the generated text includes ``Output:'', we consider the output to be the text after this token.

\paragraph{Hardware}
We ran our experiments on an instance that is equipped with NVIDIA A100 with 80 GB GPU Memory. The instance has eight 8-core CPU, each of which has access to 32 GB RAM. 

\paragraph{Models}
We used HuggingFace to load the large language models in \emph{Bfloat16 precision}. To optimize the LoRa adapters for these models, we used \emph{adamw\_bnb\_8bit,} which uses 8-bit float precision to represent the states of the optimizer (e.g., Momentum).


\begin{table}[t]
	\small
	\renewcommand{\arraystretch}{1.0}
	\setlength{\tabcolsep}{5pt}
	\centering
	\begin{tabular}{llll}
		\toprule
		\textbf{Hyperparameter} & \textbf{Low} & \textbf{High}& \textbf{Step}\\
		\midrule
		
		Learning rate & $10^{-6}$ & $10^{-4}$ & --\\
		Early stopping threshold & $10^{-3}$ & $10^{-4}$ & $10^{-4}$\\
		\bottomrule
	\end{tabular}
	\caption{The value range for the hyperparameters in the in-task and leave-one-task-out experiments. For learning rate, we did not choose any pre-defined step and sampled from all real numbers between the low and high values.}
	\label{table-hyper-parameter-range}
\end{table}


\begin{table}[t]
	\small
	\renewcommand{\arraystretch}{1.0}
	\setlength{\tabcolsep}{5pt}
	\centering
	\begin{tabular}{ll}
		\toprule
		\textbf{Hyperparameter}& \textbf{Value}\\
		\midrule
		Epochs & 30 \\
		Early stopping patience & 3  \\
		Mini Batch size & 4 \\
		Accumulation steps & 4 \\ 
		Early stoppling metric & Loss\\
		Cutoff len & 1024\\
		Lora rank & 16 \\
		Lora dropout & 0.05 \\
		Lora alpha & 16 \\
		Seed & 1517\\
		\bottomrule
	\end{tabular}
	\caption{Hyperparameters that are shared for the large language models in the in-task and leave-one-task-out experiments.}
	\label{table-hyper-parameters-shared}
\end{table}

\paragraph{Hyperparameter Optimization}
To optimize the models, we used the \emph{Tree-structured Parzen Estimator} \cite{watanabe:2023} implementation in Optuna%
\footnote{\url{https://optuna.org/}} 
with 10 trials to fine-tune the models in all experiments. We consistently used the same range of values, which are shown in Table~\ref{table-hyper-parameter-range}. Table~\ref{table-hyper-parameters-cols-bs-lr} shows the parameters with the best performance on the validation sets of the in-task and leave-one-task-out experiments. The values of other parameters that are used in the experiments are shown in Table~\ref{table-hyper-parameters-shared}. For fine-tuning on the target tasks, we always used 20 epochs, except for unit segmentation~\cite{ajjour:2017} where we used 30 epochs. For the leave-one-task-out experiment, we chose one of the validation tasks, stance classification \cite{stab:2018} to optimize the hyperparameters of the models.

\begin{table*}[t]
	\small
	\renewcommand{\arraystretch}{1.0}
	\setlength{\tabcolsep}{3pt}
	\centering
	\begin{tabular}{ll}
		\toprule
		\textbf{Model} & \textbf{Link}\\
		\midrule
		\texttt{DeepSeek-R1-7b}& \url{http://hf.co/deepseek-ai/DeepSeek-R1-Distill-Qwen-7B}\\
		\texttt{DeepSeek-R1-32b}& \url{https://hf.co/deepseek-ai/DeepSeek-R1-Distill-Qwen-32B}\\
		\addlinespace
		\texttt{Qwen3-4b}& \url{https://hf.co/Qwen/Qwen3-4B}\\
		\texttt{Qwen3-32b}& \url{https://hf.co/Qwen/Qwen3-32B}\\
		\addlinespace							   
		\texttt{Mistral-7b}& \url{https://hf.co/mistralai/Mistral-7B-Instruct-v0.3}\\
		\texttt{Mistral-22b}& \url{https://hf.co/mistralai/Mistral-Small-Instruct-2409}\\
		\texttt{Mixtral-8x7b}& \url{https://hf.co/mistralai/Mixtral-8x7B-Instruct-v0.1}\\
		\addlinespace
		\texttt{Llama-3.1-8b}& \url{https://hf.co/meta-llama/Llama-3.1-8B-Instruct}\\
		\texttt{Llama-3.3-70b}& \url{https://hf.co/meta-llama/Llama-3.3-70B-Instruct}\\
		\addlinespace
		\texttt{Phi-3.5-MoE-7.6b}&\url{https://hf.co/microsoft/Phi-mini-MoE-instruct}\\
		\addlinespace
		
		\bottomrule
	\end{tabular}	
	\caption{The urls for the hugging face models used in the paper}
	\label{table-model-links}
\end{table*}	

\begin{table*}[t]
	\small
	\renewcommand{\arraystretch}{1.0}
	\setlength{\tabcolsep}{5pt}
	\centering
	\begin{tabular}{llccc}
		\toprule
		\textbf{Task} & \textbf{Prompting Technique}  & \textbf{Top-k} & \textbf{Min Tokens} & \textbf{Max Tokens}\\
		\midrule
		
		\multirow{2}{*}{Aspect Detection} & Few-shot-prompting & 	50	 & 	32 &	1024 			\\
										  & Chain-of-thought   & 	50	 &	32 &	2048			\\
		
		\addlinespace
		Argument  							& Few-shot-prompting& 	50	 & 	32	&	2048			\\
		Unit Segmentation				  & Chain-of-thought    & 	50	 &	32	&	2048			\\
		\addlinespace
		\multirow{2}{*}{Fallacy Detection}  & Few-shot-prompting & 	50	 & 	32	&  1024				\\
										  & Chain-of-thought   & 	50	 &	32	&  2048				\\
		\addlinespace
		\multirow{2}{*}{Generation Tasks}  & Few-shot-prompting & 	50	 & 	32	& 512				\\
											& Chain-of-thought   & 	50	 & 	32	& 1024				\\
		\addlinespace
		\multirow{2}{*}{Classification Tasks}  & Few-shot-prompting & 1	 & 	16	& 	32				\\
										   & Chain-of-thought   	& 1	 & 	16	&	300				\\

		\bottomrule
	\end{tabular}
	\caption{The generation parameters used in all experiments for the segmentation tasks (Argument Unit Segmentation~\cite{habernal:2015, alkhatib:2016, stab:2017a}, Aspect Detection~\cite{schiller:2021}, and Fallacy Detection~\cite{goffredo:2023}, generation tasks, and classification tasks. Min tokens and max tokens refer to the count of tokens to be generated for each task. Classification tasks include all tasks in the benchmark whose output is a class without completion. Generation tasks cover argument generation tasks. The parameters for few-shot prompting is also used for zero-shot prompting.}
	\label{table-generation-parameters}
\end{table*}

\paragraph{Benchmark Distribution}

The following tasks from the released benchmark are available for download upon agreeing to the license. To do so, please contact the authors. \newpage

\begin{itemize}
\setlength{\itemsep}{0pt}			
	\item Suboptimal Claim Detection \cite{skitalinskaya:2023a}
	\item Claim Optimization \cite{skitalinskaya:2023b}
	\item Argument Ranking \cite{skitalinskaya:2021}
	\item Claim Improvement Suggestions ~\cite{skitalinskaya:2023a}
	\item Argument Summarization \cite{roush:2024}
\end{itemize}


\begin{table*}[!t]
	\small
	\renewcommand{\arraystretch}{1.0} 
	\setlength{\tabcolsep}{5pt}       
	\centering
	\begin{tabular}{p{7em}rrrrrrrr}
		\toprule
		& \multicolumn{2}{c}{\texttt{DeepSeek-R1-7b}} & \multicolumn{2}{c}{\texttt{DeepSeek-R1-32b}} & \multicolumn{2}{c}{\texttt{Llama-3.1-8b}} & \multicolumn{2}{c}{\texttt{Mistral-7b}} \\
		\cmidrule(lr){2-3} \cmidrule(lr){4-5} \cmidrule(lr){6-7} \cmidrule(lr){8-9}
		\textbf{Task} & \makecell{\textbf{LR}} & \textbf{EST}& \makecell{\textbf{LR}} & \makecell{\textbf{EST}} & \makecell{\textbf{LR}} & \makecell{\textbf{EST}} & \makecell{\textbf{LR}} & \makecell{\textbf{EST}} \\
		\midrule
		\multicolumn{9}{l}{\textbf{Training on the Target Task}}\\ 
		\addlinespace
		\multirow{2}{2cm}{\textbf{Unit Segmentation}}  & $4.98 \cdot 10^{-5}$ & $10^{-3}$ & $8.25 \cdot 10^{-5}$ & $10^{-4}$ & $8.06 \cdot 10^{-5}$ & $10^{-4}$  & $3.49 \cdot 10^{-5}$ & $10^{-4}$ \\
		\\
		\addlinespace

	\multirow{2}{2cm}{\textbf{Argument Similarity}}    & $4.19 \cdot 10^{-5}$ & $10^{-3}$  & $7.33 \cdot 10^{-5}$ & $7\cdot 10^{-3}$  & $3.74 \cdot 10^{-5}$ & $10^{-4}$  & $9.83 \cdot 10^{-5}$ & $7 \cdot 10^{-4}$ \\
		\\
		\addlinespace

		\multirow{2}{2cm}{\textbf{Argument Rating}}   & $9.93 \cdot 10^{-5}$ & $10^{-4}$ & $2.53 \cdot 10^{-5}$ & $10^{-4}$ & $3.28 \cdot 10^{-5}$ & $4 \cdot 10^{-4}$ & $3.49 \cdot 10^{-5}$ & $4 \cdot 10^{-4}$\\			
		\\
		\addlinespace
		
		\multirow{2}{2cm}{\textbf{Fallacy Detection}}  & $3.22 \cdot 10^{-5}$ & $10^{-3}$ & $7.71 \cdot 10^{-5}$ & $10^{-4}$ & $3.13 \cdot 10^{-5}$ & $ 8 \cdot 10^{-4}$ &  $1.31 \cdot 10^{-5}$  & $7 \cdot 10^{-4}$\\		 	
		\\
		\addlinespace
		
		\multirow{2}{2cm}{\textbf{Counterargument Generation}}  & $8.96 \cdot 10^{-5}$ & $10^{-4}$ & $7.49 \cdot 10^{-5}$ & $10^{-4}$ & $9.98 \cdot 10^{-5}$ & $10^{-4}$ & $9.88 \cdot 10^{-5}$ & $10^{-4}$\\
		\\
		\midrule
		\multicolumn{9}{l}{\textbf{Training on all other Tasks}}\\
		\addlinespace
		\multirow{2}{2cm}{\textbf{Five Target tasks}}	 & $9.51 \cdot 10^{-5}$ & $10^{-4}$ & $8.65 \cdot 10^{-5}$ & $10^{-4}$ & $3.99 \cdot 10^{-5}$ &$10^{-4}$ & $2.87 \cdot 10^{-5}$ & $10^{-4}$\\		
		\\
		\bottomrule
		\addlinespace
		 
		\addlinespace
		\addlinespace
		\addlinespace
		\addlinespace
		\toprule
		& \multicolumn{2}{c}{\texttt{Mistral-small}} & \multicolumn{2}{c}{\texttt{Qwen3-4b}} & \multicolumn{2}{c}{\texttt{Qwen3-32b}} & \multicolumn{2}{c}{\texttt{Phi-3.5-MoE-7.6b}} \\
		\cmidrule(lr){2-3} \cmidrule(lr){4-5} \cmidrule(lr){6-7} \cmidrule(lr){8-9}
		\textbf{Task} &  \makecell{\textbf{LR}} & \makecell{\textbf{EST}}  & \makecell{\textbf{LR}} & \makecell{\textbf{EST}}  & \makecell{\textbf{LR}} & \makecell{\textbf{EST}}  & \makecell{\textbf{LR}} & \makecell{\textbf{EST}} \\
		\midrule
		\multicolumn{9}{l}{\textbf{Training on the Target Task}}\\
		\addlinespace
		\multirow{2}{2cm}{\textbf{Unit Segmentation}}  & $8.06 \cdot 10^{-5}$ & $5 \cdot 10^{-3}$  & $8.6 \cdot 10^{-5}$ & $10^{-4}$  & $6.05 \cdot 10^{-5}$ & $9 \cdot 10^{-3}$  & $9.79 \cdot 10^{-5}$ &  $10^{-3}$\\
				\\
				\addlinespace

		\multirow{2}{2cm}{\textbf{Argument Similarity}}  & $5.36 \cdot 10^{-5}$ & $5 \cdot 10^{-4}$ & $6.08\cdot 10^{-5}$ & $10^{-3}$  & $2.86 \cdot 10^{-5}$ & $5\cdot10^{-4}$  & $6.72 \cdot 10^{-5}$ & $ 10^{-3}$\\
				\\
				\addlinespace

		\multirow{2}{2cm}{\textbf{Argument Rating}} 			 & $1.76 \cdot 10^{-5}$ & $8 \cdot 10^{-4}$ & $1.94 \cdot 10^{-5}$ & $10^{-3}$& $7.09 \cdot 10^{-5}$ & $ 10^{-4}$ & $6.99 \cdot 10^{-5}$ & $6 \cdot 10^{-4}$\\
				\\
				\addlinespace
				
		\multirow{2}{2cm}{\textbf{Fallacy Detection}}			 & $6.48 \cdot 10^{-5}$ & $ 10^{-4}$ & $2.98 \cdot 10^{-5}$ & $ 10^{-3}$ & $6.40 \cdot 10^{-5}$ & $7 \cdot 10^{-4}$ & $7.57 \cdot 10^{-5}$ & $ 10^{-4}$\\
				\\
				\addlinespace
				
		\multirow{2}{2cm}{\textbf{Counterargument Generation}} 	& $6.53 \cdot 10^{-5}$ & $ 10^{-4}$  & $9.86 \cdot 10^{-5}$ & $4 \cdot 10^{-4}$ & $8.70 \cdot 10^{-5}$ & $6 \cdot 10^{-4}$ & $8.55 \cdot 10^{-5}$ & $\cdot 10^{-4}$\\
				\\
				\midrule
				
		\multicolumn{9}{l}{\textbf{Training on all other Tasks}}\\
		
		\addlinespace
		\multirow{2}{2cm}{\textbf{Five Target tasks}}  & $7.06 \cdot 10^{-5}$ & $ 10^{-4}$ & $9.79 \cdot 10^{-5}$ & $ 10^{-4}$ & $3.19 \cdot 10^{-5}$ & $ 10^{-4}$ & $5.88 \cdot 10^{-5}$ & $ 10^{-4}$\\
		\\
		\bottomrule
	\end{tabular}
	\caption{The hyperparameters learning rate (LR) and early stopping threshold (EST) when \emph{training on the target task} and \emph{training on all other tasks}. The target tasks are: unit segmentation~\cite{habernal:2015}, argument rating~\cite{wachsmuth:2017b}, argument similarity~\cite{reimers:2019}, fallacy detection~\cite{habernal:2018b}, and counterargument generation~\cite{hua:2018}.}
	\label{table-hyper-parameters-cols-bs-lr} 
	\end{table*}

\subsection{Experiments}
\label{sec:extra-experiments}

\subsubsection{Manual evaluation}
\label{manual-evaluation-instructions}
\begin{figure*}
	\centering\small
	\includegraphics[width=.9\textwidth]{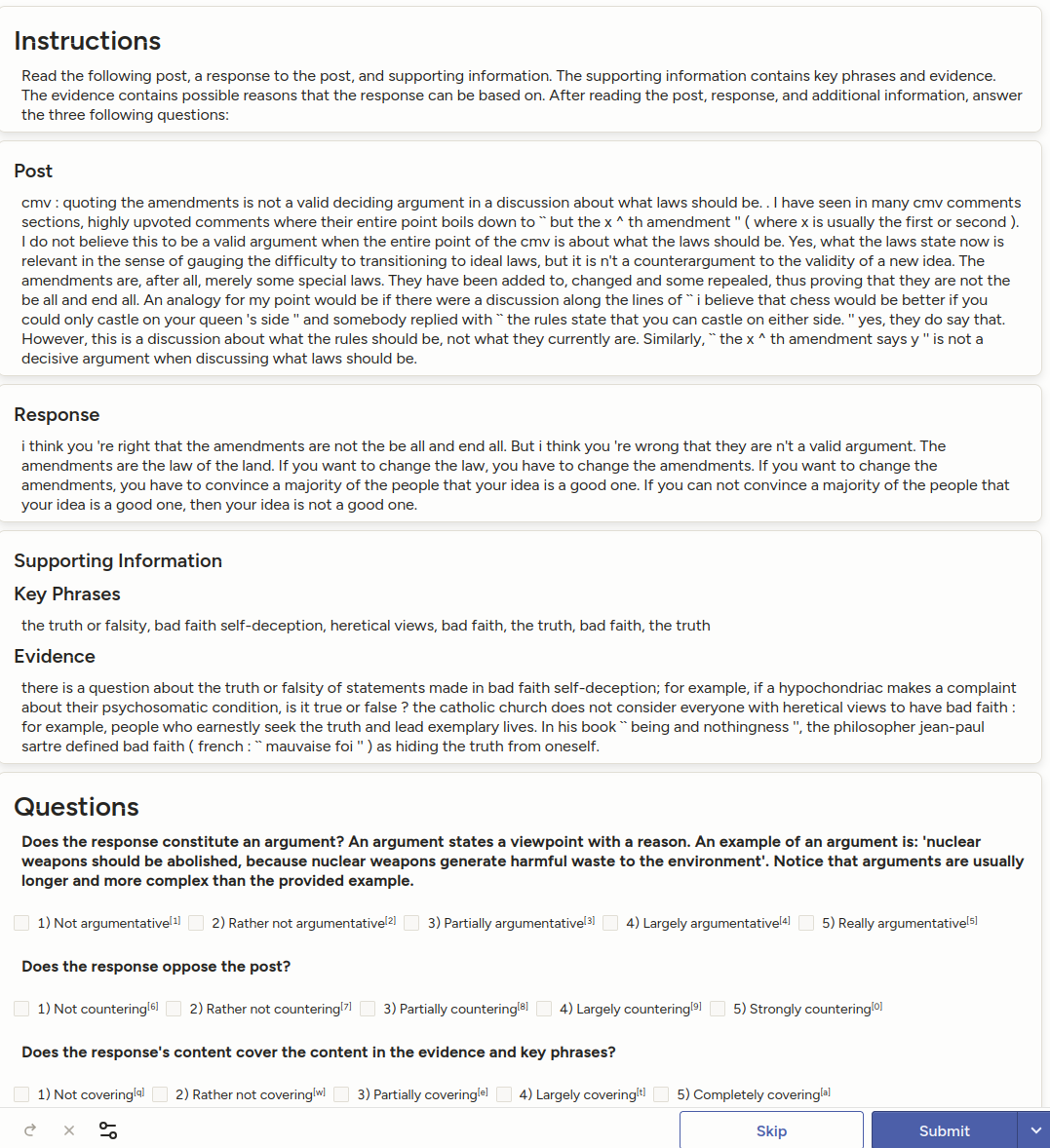}
	\caption{The interface for the manual evaluation of counterarguments that are generated by \texttt{Mistral-22b}, \texttt{Mistral-7b}, \texttt{Qwen3-32b}, and \texttt{Phi-3.5-MoE-7.6b} on the counterargument generation task~\cite{hua:2018}.}
	\label{fig:annotation-interface}
\end{figure*}

The manual evaluation started with a pilot study where two authors labeled 50 counterarguments for five randomly sampled arguments from counterargument generation~\cite{hua:2018}. The goal of the pilot study was to develop the annotation guideline and validate the counterargument collection process. For the main study, which we report on in the main body, we recruited three annotators from three different countries on Upwork: the USA, Canada, and India. We have paid each annotator 15 dollars per hour and requested from them a confirmation that they are fine with labeling sensitive content about social issues like racism. All in all, the crowdworkers spend 14 hours on the annotation task. A screenshot of the interface and the instructions are provided in Figure~\ref{fig:annotation-interface}. Detailed definitions of the quality criteria on which the annotators assessed the counterarguments are listed here:

\begin{itemize}
	\item \textbf{Argumentativeness}: whether the counterargument constitutes an argument. 
	\item \textbf{Countering}: whether the counterargument takes an opposite stance to the argument.
	\item \textbf{Relevance}: whether the counterargument's content covers the content of the evidence and key phrases.
\end{itemize}

\end{document}